\PassOptionsToPackage{table}{xcolor}
\documentclass{article}

\usepackage[preprint]{neurips_2026}
\usepackage[utf8]{inputenc} 
\usepackage[T1]{fontenc}    
\usepackage{hyperref}       
\usepackage{url}            
\usepackage{booktabs}       
\usepackage{amsfonts}       
\usepackage{nicefrac}       
\usepackage{microtype}      
\usepackage{xcolor}         
\usepackage{amsmath} 
\usepackage{enumitem}
\usepackage{adjustbox}
\usepackage{graphicx}
\usepackage{colortbl}
\usepackage{tabularx}
\usepackage{makecell}
\usepackage{booktabs}
\usepackage{multirow}
\usepackage{array}
\usepackage{newtxtext,newtxmath}
\usepackage{float}
\usepackage{wrapfig}
\definecolor{myred}{RGB}{200, 40, 40}
\definecolor{mygreen}{RGB}{40, 150, 60}
\definecolor{baselineblue}{RGB}{235,243,251}     
\definecolor{withoutintblue}{RGB}{221,236,248}   
\definecolor{oursbrightblue}{RGB}{188,222,250}   
\usepackage{titletoc}
\usepackage[most]{tcolorbox}
\usepackage{enumitem}

\newcommand{\best}[1]{\textbf{#1}}
\newcommand{\second}[1]{\underline{#1}}

\newcommand{\metricmain}[1]{{\fontsize{12.6}{13.2}\selectfont #1}}

\newcommand{\upann}[1]{%
  \textcolor[rgb]{0.93,0.32,0.32}{%
    \raisebox{0.15ex}{\fontsize{6.5}{6.0}\selectfont($\uparrow$\kern-0.05em #1)}%
  }%
}

\newcommand{\downann}[1]{%
  \textcolor[rgb]{0.18,0.62,0.34}{%
    \raisebox{0.15ex}{\fontsize{6.5}{6.0}\selectfont($\downarrow$\kern-0.05em #1)}%
  }%
}

\newcommand{\sameann}{%
  \textcolor[rgb]{0.93,0.32,0.32}{%
    \raisebox{0.15ex}{\fontsize{6.5}{6.0}\selectfont($\uparrow$\kern-0.05em 0.00)}%
  }%
}

\newcommand{\metric}[2]{%
  \metricmain{#1}\hspace{0.06em}#2%
}

\definecolor{sectiongray}{RGB}{238,238,238}
\definecolor{oursblue}{RGB}{214,238,250}

\newcommand{\groupheader}[2]{%
  \rowcolor{sectiongray}
  \multicolumn{1}{l|}{} & \multicolumn{8}{c}{\textbf{#2}} \\
}
\newcommand{\grouprow}{}

\newcommand{\bluegroupheader}[2]{%
  \rowcolor{sectiongray}
  \multicolumn{1}{l|}{} & \multicolumn{8}{c}{\textbf{#2}} \\
}
\newcommand{\bluerow}{}

\newcommand{\yellowgroupheader}[2]{%
  \rowcolor{sectiongray}
  \multicolumn{1}{l|}{} & \multicolumn{8}{c}{\textbf{#2}} \\
}
\newcommand{\lightyellowrowstart}{}

\newcommand{\baselinebluerowstart}{\rowcolor{baselineblue}}
\newcommand{\withoutintbluerowstart}{\rowcolor{withoutintblue}}
\newcommand{\oursbluerowstart}{\rowcolor{oursbrightblue}}

\title{
  \texorpdfstring{
    \begin{tabular}{@{} m{2.8em} @{} m{0.95\textwidth} @{}} 
      \includegraphics[width=2.5em]{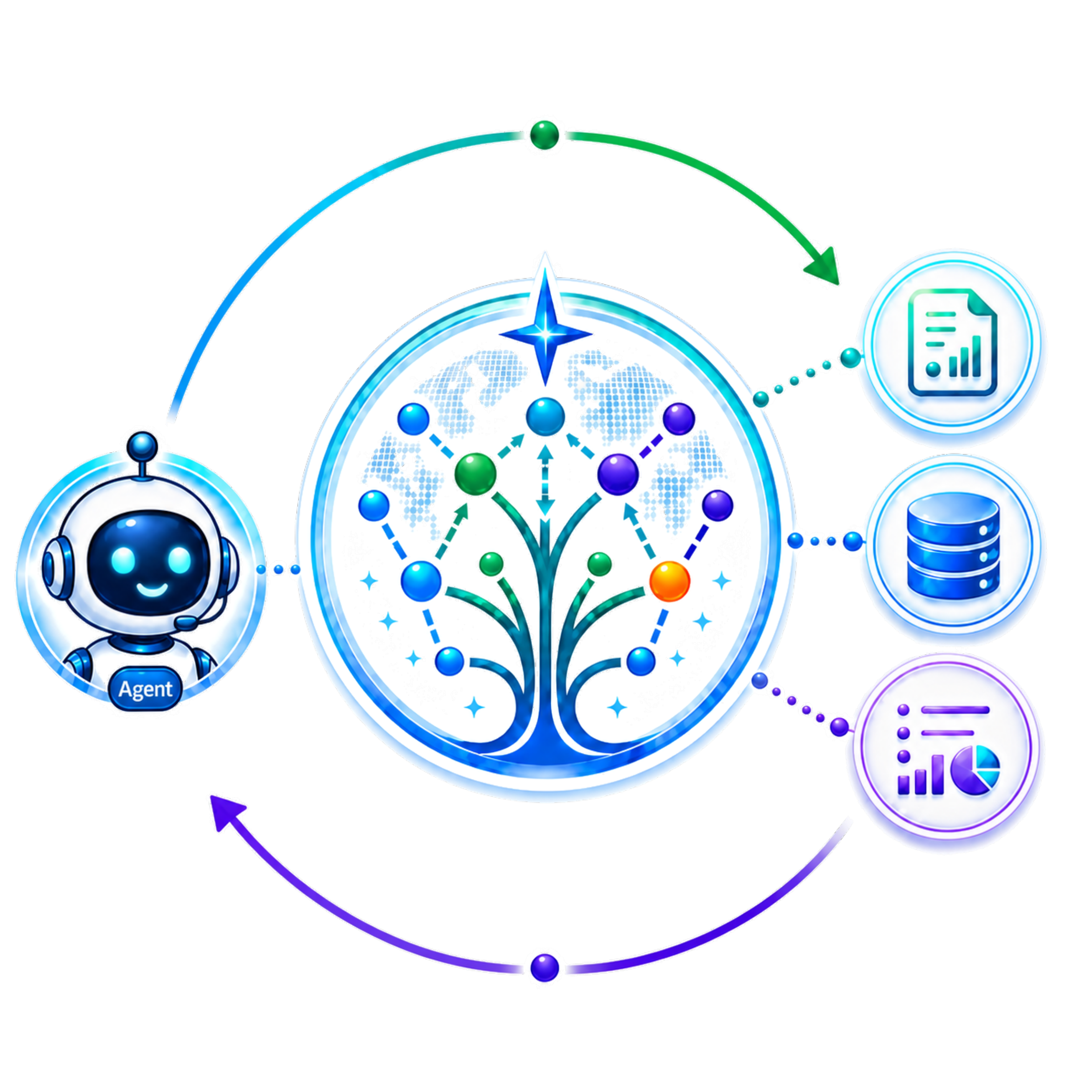} & 
      {\LARGE \raggedright \textbf{ANDES: Agent Native Data Evolving Synthesis\\ Tool for Autonomous Instruction Alignment}}
    \end{tabular}
  }{ANDES: Agent Native Data Evolving Synthesis Tool for Autonomous Instruction Alignment}
}
\author{
\textbf{Zhengyang Zhao}\textsuperscript{*1},
\textbf{Shengjie Ye}\textsuperscript{*2},
\textbf{Lu Ma}\textsuperscript{1},
\textbf{Hao Liang}\textsuperscript{1},
\textbf{Hengyi Feng}\textsuperscript{1},
\textbf{Wentao Zhang}\textsuperscript{\ensuremath{\dagger}1} \\
\textsuperscript{1} Peking University
\quad
\textsuperscript{2} Sichuan University, Chengdu\\
\texttt{zhengyangzhao25@stu.pku.edu.cn},
\texttt{yeshengjie@stu.scu.edu.cn} \\
\texttt{wentao.zhang@pku.edu.cn}
}
\begin{document}

\maketitle
\begingroup
\renewcommand{\thefootnote}{*}
\footnotetext{Equal contribution.}
\renewcommand{\thefootnote}{\ensuremath{\dagger}}
\footnotetext{Corresponding author.}
\endgroup

\begin{abstract}
AI agents are increasingly being tasked with automating AI research itself, particularly the critical post-training phase that transforms base LLMs into aligned assistants. However, recent evaluations reveal that even frontier agents struggle to perform this task. While the success of post-training fundamentally relies on acquiring high-quality data, relying on agents to autonomously curate targeted training datasets from the open web introduces severe challenges. Executing the long-horizon tasks of searching, filtering, and balancing data within noisy web environments frequently overwhelms an agent's limited context, ultimately leading to degraded dataset quality and suboptimal downstream training performance. To bridge this gap, we introduce \textsc{Andes} (Agent Native Data Evolving Synthesis), a framework that reimagines data generation as a plug-and-play \emph{agent skill}. Rather than forcing agents to devise complex data-gathering strategies from scratch, \textsc{Andes} provides an intelligent abstraction layer. By leveraging a self-evolving World Tree routing mechanism and actionable diagnostic reports, it allows trainer agents to dynamically steer data synthesis through an interactive, closed-loop interface. We demonstrate that under strict compute constraints, equipping foundationally weaker agents with \textsc{Andes} improves automated alignment, securing state-of-the-art performance on PostTrainBench and robust cross-task generalization. Our project is available at \url{https://github.com/zzy1127/ANDES}.
\end{abstract}

\vspace{-0.8em}

\begin{figure}[!h]
  \centering
  \vspace{-0.5em}
  \includegraphics[width=0.92\textwidth]{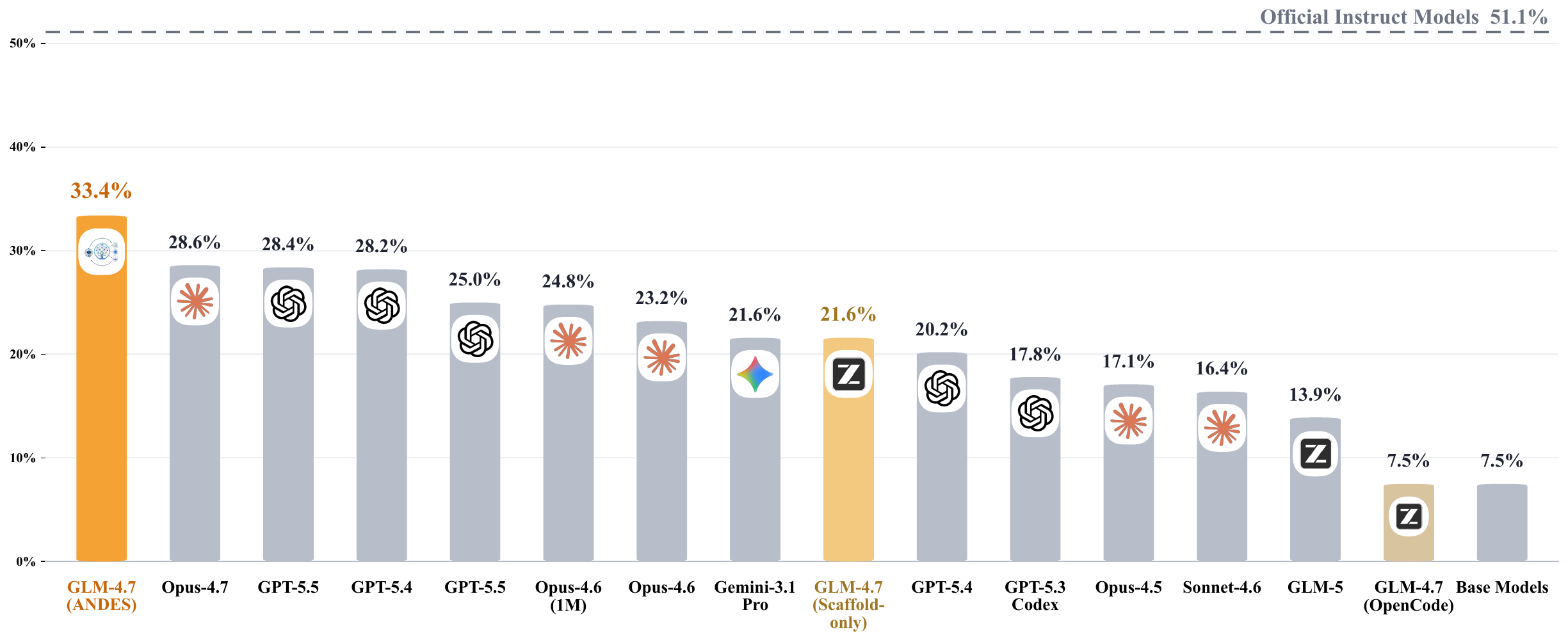}
  \vspace{-0.8em}
  \caption{\textbf{\textsc{Andes} achieves SOTA performance on PostTrainBench. }Compared to the bare execution baseline GLM-4.7 (Scaffold-only), \textsc{Andes} drives a definitive alignment leap to 33.4\%, outperforming Opus-4.7 by 4.8\%.}
  \label{fig:posttrainbench_leaderboard}
  \vspace{-1.0em}
\end{figure}

\section{Introduction}
\label{sec:introduction}



Driven by the rapid advancements in agentic capabilities, current research trends are increasingly focusing on applying autonomous agents to orchestrate the Large Language Model (LLM) post-training process\cite{posttrainbench_2026}---encompassing techniques such as SFT and RL\cite{shao2024deepseekmathpushinglimitsmathematical, wu2026generalizationsftreinforcementlearning, yu2025dapoopensourcellmreinforcement, zhao2026giftreconcilingposttrainingobjectives}. By delegating the complex decision-making of model alignment to an agent, researchers aim to replace tedious manual trial-and-error with continuous, self-correcting workflows\cite{ma2026trexautomatingllmfinetuning, jiang2025aideaidrivenexplorationspace, li2023traineragentcustomizableefficientmodel, li2026autosotaendtoendautomatedresearch}. In this context, post-training agents are expected to autonomously diagnose capability deficits and actively acquire targeted training signals to patch these weaknesses. Unlike traditional static training pipelines, this dynamic evolution requires the agent to continuously intervene with tailored data scenarios. Consequently, the efficiency and precision with which an agent can procure high-quality, diverse data fundamentally dictate the upper bound of the entire training loop.

However, existing data acquisition strategies present a severe capability trap for post-training agents. On one hand, frameworks that rely on external environments, such as open web search, demand profound foundational capabilities---including complex long-horizon planning, raw HTML parsing, and zero-shot noise filtering\cite{zhou2024webarenarealisticwebenvironment, feng2026longclibenchpreliminarybenchmarkstudy}. This imposes a prohibitively high barrier, causing weak-to-medium models to fail catastrophically before they can form a functional closed training loop. As for advanced agents equipped with comprehensive web information-gathering capabilities—such as Opus 4.7—they often remain hindered by the organization and balancing of data from disparate sources, as well as by the control of data quality; this, in turn, leads to suboptimal post-training performance. On the other hand, conventional synthetic data pipelines are predominantly implemented as static, one-shot scripts. They operate in isolation from the agent's active training cycle, stripping the agent of the ability to dynamically steer the generation direction based on real-time diagnostic feedback\cite{xu2025wizardlmempoweringlargepretrained, liang2025dataflowllmdrivenframeworkunified}. Therefore, once the agent encounters a new downstream task scenario, these methods are very likely to fail due to data domain mismatch. This disconnect inevitably leads to diversity collapse, wasted compute, and severe over-fitting to surface-level benchmark distributions. In short, there is a critical absence of an accessible, dynamically interactive high quality data synthesis interface tailored for post-training agents.

To fill this void, we introduce \textsc{Andes} (Agent Native Data Evolving Synthesis), the first agent-native data synthesis framework. Rather than forcing agents to navigate the chaotic open web or execute rigid offline scripts, \textsc{Andes} elegantly encapsulates the complexities of high-fidelity data generation into a plug-and-play \emph{agent skill}. By exposing a simple tool-calling interface, it empowers even foundationally weak agents to orchestrate sophisticated data synthesis. The trainer agent simply declares its target cognitive capability according to the downstream benchmarks, and \textsc{Andes} handles the underlying scenario abstraction, returning a highly aligned data batch alongside a real-time diagnostic report to provide guidance for adjustments in the next invocation. Under the hood, this steerable process is driven by a self-evolving World Tree routing mechanism that autonomously expands to provide diverse contexts. Coupled with the report-driven feedback loop, \textsc{Andes} transforms data generation into a dynamic, closed-loop game, allowing the agent to actively filter data, seamlessly calibrate its configurations for subsequent invocations, and proactively steer the trajectory away from logical repetition.

The main contributions of this work are summarized as follows:
\begin{itemize}
\item We introduce a new perspective on agentic post-training, reconceptualizing data synthesis not as a static, offline pipeline, but as a plug-and-play \emph{agent skill}. This conceptual shift significantly lowers the capability threshold, enabling agents to autonomously acquire targeted, high-quality training data.
\item We implement \textsc{Andes}, a plug-and-play synthesis tool for trainer agents. Powered by self-evolving World Tree routing and report-driven feedback, \textsc{Andes} abstracts away the intricate mechanics of data curation, allowing the trainer agent to focus entirely on high-level optimization strategies and therefore enables precise control over data generation to acquire high-quality, diverse training data.
\item We comprehensively evaluate \textsc{Andes} across diverse benchmarks, demonstrating that its integration into the active post-training loop enables foundationally weaker models to achieve remarkable autonomous self-improvement and secure state-of-the-art performance. Furthermore, extensive ablation and analytical studies validate the efficacy of our core designs.
\end{itemize}

\section{Related Works}
\label{sec:related_works}

\subsection{SFT Data Synthesis}

The growing demand for high-quality supervised fine-tuning (SFT) data has driven extensive research into synthetic data generation. Early works bootstrap instruction data from seed prompts (Self-Instruct~\cite{wang2023selfinstructaligninglanguagemodels}, WizardLM~\cite{xu2025wizardlmempoweringlargepretrained}, Condor~\cite{cao2025condorenhancellmalignment}), but often suffer from limited seed diversity. To reduce explicit seed dependence, Magpie~\cite{xu2024magpiealignmentdatasynthesis} and MATRIX-Gen~\cite{wang2025matrixpeertopeermultiagentsynthetic} explore template-based and multi-agent generation, though trading flexibility for higher orchestration costs. Another thread focuses on structured synthesis and quality verification via base-refine paradigms (BARE~\cite{zhu2025bareleveragingbaselanguage}), executable feedback (AutoIF~\cite{dong2024selfplayexecutionfeedbackimproving}), and prompt decomposition (UltraIF~\cite{an2025ultraifadvancinginstructionfollowing}, DecIF~\cite{hui2025decifimprovinginstructionfollowingmetadecomposition}). More recently, to adapt to evolving model capabilities, Middo~\cite{tang2025middomodelinformeddynamicdata} and WIST~\cite{li2026wistwebgroundediterativeselfplay} introduced dynamic, closed-loop synthesis frameworks. Despite these advances, existing approaches remain fundamentally human-centric or rely on rigid offline workflows. They critically lack the standardized, dynamically interactive tool interfaces required for an autonomous agent to plug-and-play and proactively steer the data acquisition process.

\subsection{Automated Research}

Automating the full research pipeline has emerged as a central challenge in AI development. Recent works like AIDE~\cite{jiang2025aideaidrivenexplorationspace}, AlphaEvolve~\cite{novikov2025alphaevolvecodingagentscientific}, AUTO-SOTA~\cite{li2026autosotaendtoendautomatedresearch}, and the AI Scientist series~\cite{lu2024aiscientistfullyautomated,yamada2025aiscientistv2workshoplevelautomated} demonstrate the potential of agents to autonomously iterate on experiments and produce algorithmic discoveries. To systematically evaluate these capabilities, benchmarks such as MLE-bench~\cite{chan2025mlebenchevaluatingmachinelearning} and MLAgentBench~\cite{huang2024mlagentbenchevaluatinglanguageagents} provide diverse end-to-end ML tasks for agent assessment. Focusing specifically on the post-training phase, TREK~\cite{ma2026trexautomatingllmfinetuning} automates the training lifecycle via tree-based exploration, while PostTrainBench~\cite{posttrainbench_2026} reveals that frontier agents still lag significantly behind official instruction-tuned models under bounded compute budgets. Recognizing that the primary bottleneck in these workflows lies in the complex curation of high-quality training data, we introduce \textsc{Andes} as a dedicated, agent-native synthesis tool to narrow this capability gap and enhance autonomous post-training pipelines.
\section{Methodology}
\label{sec:methodology}

\subsection{Overview}
\label{subsec:overview}
\begin{figure}[htbp]
    \centering
    \includegraphics[width=\textwidth]{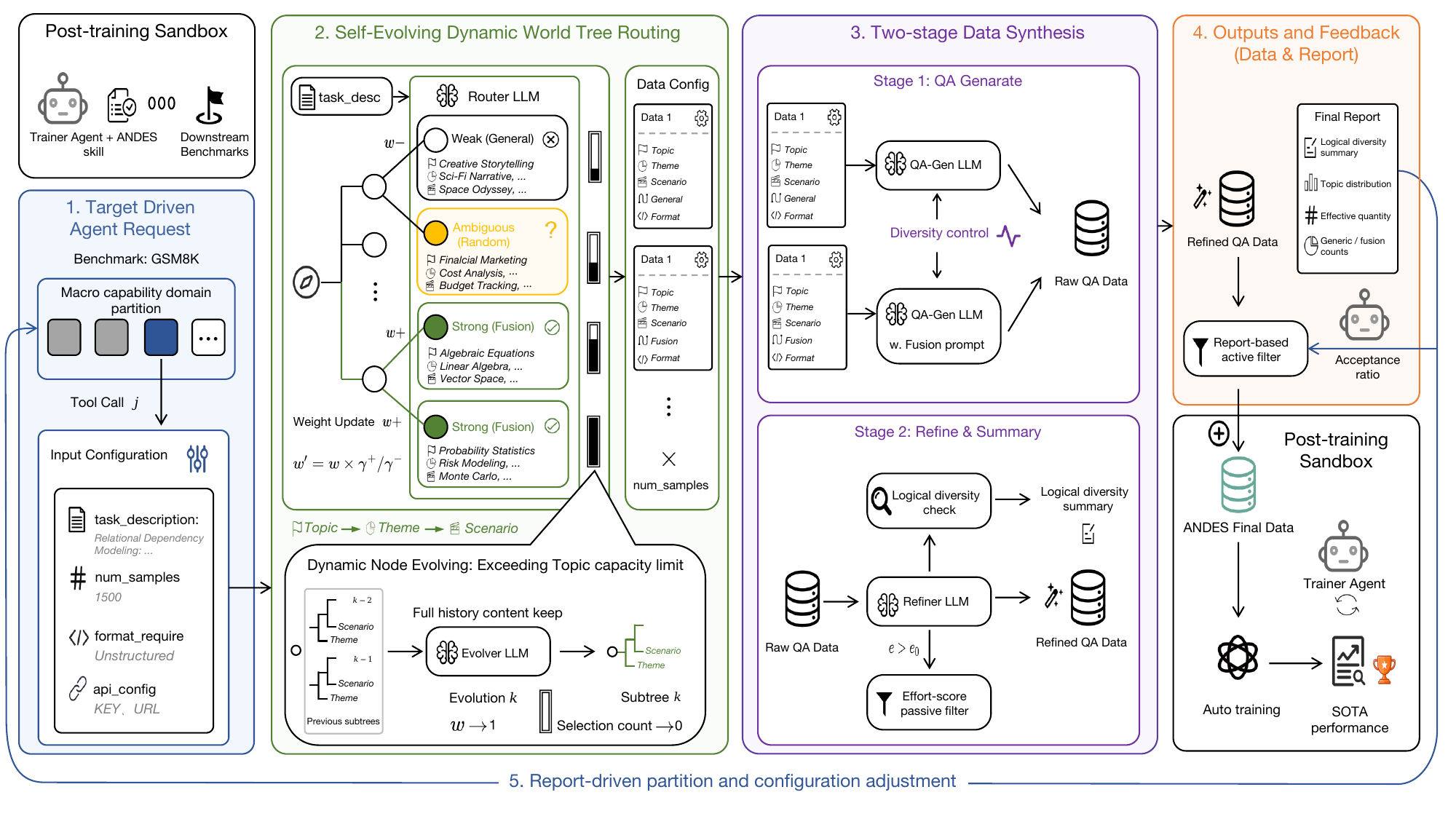}
    \caption{\textbf{Overview of \textsc{Andes}.} Guided by the \textsc{Andes} skill, a trainer agent decomposes downstream benchmarks into capability domains and invokes \textsc{Andes} once per domain; each call routes sampled topics through a self-evolving world tree, runs a two-stage QA generation and refinement pipeline, and returns a refined dataset together with a synthesis report that drives the trainer agent's configuration of the next call.}
    \label{fig:overview}
\end{figure}

We introduce \textsc{Andes}, an Agent-Native Data-Evolving Synthesis tool designed to be invoked as a plug-and-play module inside a post-training agent loop. As shown in Figure~\ref{fig:overview}, guided by the \textsc{Andes} skill, the trainer agent first abstracts the objective of downstream benchmarks into a set of capability dimensions and calls \textsc{Andes} once per dimension.
The $j$-th call takes a task description $\tau_j$, a sample budget $N_j$, and an optional format protocol $\phi_j$ as input configuration, and returns a refined dataset $\mathcal{D}_j$ together with a synthesis report $\mathcal{R}_j$.
The trainer agent reads $\mathcal{R}_j$ to adjust the configuration of the next call, forming a closed loop over multiple rounds.
This section details four core stages of the proposed framework: target-driven agent request, self-evolving dynamic world tree routing, two-stage data synthesis, and report-driven partition and configuration adjustment.

\subsection{Target-Driven Agent Request}
\label{subsec:domain_partitioning}

Before invoking \textsc{Andes}, the trainer agent analyzes the target post-training objective benchmarks and decomposes them into $K$ macro capability domains $\{d_1, \ldots, d_K\}$.
To avoid overfitting the synthesis trajectory to the surface patterns of any particular evaluation target, we embed this decomposition strategy into the \textsc{Andes} skill: rather than directly enumerating task-specific requirements, the trainer agent is instructed to first abstract upward to the underlying capability dimensions and then partition along those dimensions.
This ensures that the generated data covers transferable capabilities rather than narrow task-specific patterns, maintaining diversity across synthesis rounds.
For each domain $d_j$, the trainer agent constructs a task description $\tau_j$ that characterizes the capability focus, allocates a sample budget $N_j$, and selects a format protocol $\phi_j$ as the skill guided.
The triple $(\tau_j, N_j, \phi_j)$ then forms the input configuration of the $j$-th \textsc{Andes} tool call.

\subsection{Self-Evolving Dynamic World Tree Routing}
\label{subsec:world_tree}

To facilitate precise task-specific grounding, \textsc{Andes} incorporates a self-evolving world tree $\mathcal{W}$ structured as $\text{Topic} \to \text{Theme} \to \text{Scenario}$ (e.g.\ Cybersecurity $\to$ Information Leak Prevention $\to$ Prevention Measures), covering more than $1{,}000$ scenarios at initialization.
Let $\mathcal{T} = \{t_1, t_2, \ldots, t_M\}$ denote the set of top-level topics.
Each topic $t \in \mathcal{T}$ is associated with a sampling weight $w_t > 0$ and a selection count $c_t$, both uniformly initialized at the start of a synthesis session.

\subsubsection{Weighted Sampling and Suitability Routing}
\label{subsubsec:routing}

Given task description $\tau$ and batch size $B$, we sample $B$ topics proportionally to their current weights:
\[
t_i \sim \mathrm{Categorical}(w_{t_1}, \ldots, w_{t_M}), \quad i = 1, \ldots, B.
\]
For each sampled topic $t_i$, a theme $h_i$ and a corresponding scenario $s_i$ are drawn uniformly from its subtree.
A Router LLM then evaluates all $B$ scenarios in a single pass and classifies each as \textit{Strong}, \textit{Ambiguous}, or \textit{Weak} with respect to $\tau$.
Scenarios rated \textit{Strong} are sent to the \emph{fusion} generation track and those rated \textit{Weak} to the \emph{generic} track, while \textit{Ambiguous} scenarios are stochastically assigned to fusion to balance task alignment and diversity.
This yields a binary fusion flag $f_i \in \{0, 1\}$ for each sample.
The flag, together with $(t_i, h_i, s_i)$, the task description $\tau$, and the format protocol $\phi$, constitutes a routing configuration
\[
r_i = (t_i,\, h_i,\, s_i,\, f_i,\, \tau,\, \phi),
\]
which is dispatched to the subsequent data synthesis pipeline.

After routing, the topic weights are updated based on the suitability ratings:
\begin{equation}
w_t \leftarrow \begin{cases}
  \gamma^{+} \cdot w_t & \text{topic } t \text{ has a \textit{Strong}-rated sample}, \\
  \max(\epsilon,\; \gamma^{-} \cdot w_t) & \text{topic } t \text{ has a \textit{Weak}-rated sample},
\end{cases}
\label{eq:weight_update}
\end{equation}
where $\gamma^{+} > 1$, $\gamma^{-} \in (0,1)$, and $\epsilon > 0$ is a floor that keeps every topic reachable.
This update gradually shifts the synthesis budget toward topics semantically aligned with $\tau$ while preserving residual coverage over less relevant ones.

\subsubsection{Dynamic Node Evolution}
\label{subsubsec:node_evolution}
The weighting scheme gradually concentrates sampling on a small set of task-aligned topics, which would otherwise exhaust their finite scenario pools and force the synthesis to keep drawing from the same nodes.
To keep the taxonomy fresh under this pressure, we trigger a subtree evolution for topic $t$ once its selection count $c_t$ exceeds a saturation threshold proportional to the current size of its subtree:
\[
c_t > \rho \cdot |\mathcal{S}_t|,
\]
where $|\mathcal{S}_t|$ denotes the number of scenarios currently under topic $t$ and $\rho \in (0,1)$ controls how aggressively the topic is expanded.
Let $k$ denote the index of the current evolution event for topic $t$ (i.e., the $k$-th time $t$ triggers evolution).
An Evolver LLM is invoked to produce a new theme--scenario subtree, conditioned on the full record of previously generated subtrees for the same topic:
\[
\text{Subtree}_k = \mathrm{Evolver}\!\left(t,\; \tau,\; \{\text{Subtree}_1, \ldots, \text{Subtree}_{k-1}\}\right),
\]
so that each expansion is aware of what has already been added and actively avoids redundancy across evolution rounds.
The new nodes are then written back into $\mathcal{W}$, $c_t$ is reset to zero, and $w_t$ is restored to its initial value to prevent the just-expanded topic from being immediately over-sampled and triggering another expansion.

\subsection{Two-Stage Data Synthesis}
\label{subsec:two_stage_synthesis}

Each routing configuration $r_i$ is processed by a two-stage pipeline: Stage~1 generates raw QA pairs, and Stage~2 refines these pairs and produces the synthesis report.

\subsubsection{Stage 1: QA Generation}
\label{subsubsec:generation}

Stage~1 turns each routing configuration $r_i$ into a concrete QA sample through two sequential LLM calls: the first synthesizes a question $x_i$ and the second produces its answer $y_i$, together yielding raw QA data $\{(x_i, y_i)\}$.
The construction of the question-generation prompt is governed by the fusion flag $f_i$.
When $f_i = 1$ (fusion track), $\tau$ is injected as a hard constraint: the sampled scenario $s_i$ acts as a contextual wrapper that diversifies the question surface, while the underlying logic is required to exercise the capability described by $\tau$.
To further promote intra-batch diversity, questions are generated with explicit control over auxiliary dimensions such as difficulty level and narrative form.
When $f_i = 0$ (generic track), the prompt conditions only on $h_i$ and $s_i$ without any task-specific constraint, producing general-purpose instruction data in the spirit of knowledge-tree-based synthesis~\cite{cao2025condorenhancellmalignment}.
If a format protocol $\phi$ is specified, it is attached to the answer-generation prompt of fusion-track samples so that the response obeys the required output structure.

\subsubsection{Stage 2: Refine and Summary}
\label{subsubsec:refinement}
The raw outputs from Stage~1 still suffer from two complementary forms of imperfection: individual responses may lack of reasoning depth, and a batch data may collapse onto a small set of repeated reasoning templates.
Stage~2 addresses these two issues in turn, and packages the resulting evidence into the synthesis report.

Raw QA data $\{(x_i, y_i)\}$ is first passed to a Refiner LLM, which critiques each response and produces structured feedback covering strengths, weaknesses, and a revision suggestion, together with an effort score $e_i \in \{1, \ldots, 5\}$ that estimates the cost of repairing the response.
Samples with $e_i > e_0$ are discarded by an effort-score passive filter before any rewriting, since heavily defective responses are unlikely to yield reliable training signal.
The retained responses are rewritten conditioned on the original QA pair and the critique, producing refined QA data $\{(x_i, \hat{y}_i)\}$.

We further perform a logical diversity check to suppress reasoning-template repetition among fusion-track samples.
Several mini-batches are sampled from the fusion-track subset of the refined data and evaluated by an LLM, which labels each batch as \textit{Low}, \textit{Medium}, or \textit{High} collapse---where collapse refers to cosmetically varied scenarios sharing an identical underlying reasoning template.
We restrict this check to fusion samples because they carry task-specific structure and are most susceptible to template repetition; generic-track samples are excluded by design.
Batch verdicts are merged into a short natural-language summary of the most common repeating patterns.
This diversity summary, together with the topic distribution, effective sample count, and generic/fusion counts, is compiled into the synthesis report $\mathcal{R}_j$ for the current tool call.

\subsection{Report-Driven Partition and Configuration Adjustment}
\label{subsec:cyclic_feedback}
After the $j$-th \textsc{Andes} call, the trainer agent receives $(\mathcal{D}_j, \mathcal{R}_j)$ and uses the report to close the feedback loop before the next invocation.

The trainer agent first applies a report-based active filter to $\mathcal{D}_j$. Informed by the collapse intensity recorded in $\mathcal{R}_j$, it retains a subset $\mathcal{D}_j^{\text{keep}} \subseteq \mathcal{D}_j$, applying more stringent pruning when severe logical repetition is detected.
The retained data is accumulated into the training pool $\mathcal{D} \leftarrow \mathcal{D} \cup \mathcal{D}_j^{\text{keep}}$.

Guided by \textsc{Andes} skill, the same report is then re-read as a diagnostic signal for the next invocation: the collapse pattern summary guides the trainer agent to revise $\tau_{j+1}$ toward underexplored capability angles, while the sample deficit $N_j - |\mathcal{D}_j^{\text{keep}}|$ informs an upward adjustment of $N_{j+1}$ to compensate for lost samples.
The configuration of the next call is therefore decided by the current synthesis outcome:
\[
(\tau_{j+1},\, N_{j+1},\, \phi_{j+1}) = \mathcal{F}\!\left(\tau_j,\, \phi_j,\, \mathcal{R}_j,\, N_j - |\mathcal{D}_j^{\text{keep}}|\right),
\]
where $\mathcal{F}$ denotes the trainer agent's configuration update policy conditioned on the diagnostic report and the remaining sample deficit.
Through this loop, \textsc{Andes} functions not as a one-shot generator but as an interactive synthesis substrate that the trainer agent progressively steers toward better capability coverage and data diversity.

\begin{table*}[t]
\centering
\setlength{\tabcolsep}{3.9pt}
\renewcommand{\arraystretch}{1.18}
\caption{Results of the 28-experiments on PostTrainBench (accuracy in \%). The base models include Qwen3-1.7B, Qwen3-4B, SmolLM-3B, and Gemma3-4B. The best and second-best results are highlighted in \textbf{bold} and \underline{underlined}. Numbers after arrows denote absolute changes over the corresponding zero-shot base model, with improvements in \textcolor{myred}{red} and declines in \textcolor{mygreen}{green}.}
\vspace{4pt}
\label{tab:leaderboard_results}

\begin{adjustbox}{width=\textwidth,center}
\fontsize{12.3}{14.2}\selectfont
\begin{tabular}{l|cccccccc}
\toprule
\addlinespace[0.12em]
\textbf{Method} & \textbf{AIME 2025} & \textbf{ArenaHard} & \textbf{BFCL} & \textbf{GPQA Main} & \textbf{GSM8K} & \textbf{HealthBench} & \textbf{HumanEval} & \textbf{AVG (\%)} \\[0.16em]
\midrule

\groupheader{9}{(I) Evaluation with Official Instruct Models}
\grouprow
\textbf{Official Instruct Models}
& 29.17 & 70.21 & 85.00 & 36.21 & 87.00 & 43.32 & 71.49 & 51.14 \\
\midrule

\groupheader{9}{(II) Evaluation with Base Models (Zero-Shot)}
\grouprow
\textbf{Base Models (Average)}
& 1.67 & 1.26 & 1.50 & 8.48 & 20.43 & 9.49 & 12.81 & 7.53 \\
\midrule

\bluegroupheader{9}{(III) Evaluation with Agent-Post-Trained Base Models}

\bluerow
Kimi-K2-Thinking
& \metric{1.67\textsuperscript{*}}{\sameann}
& \metric{1.26\textsuperscript{*}}{\sameann}
& \metric{1.50\textsuperscript{*}}{\sameann}
& \metric{8.48\textsuperscript{*}}{\sameann}
& \metric{14.84}{\downann{5.59}}
& \metric{9.49\textsuperscript{$\dagger$}}{\sameann}
& \metric{15.09\textsuperscript{*}}{\upann{2.28}}
& \metric{7.25}{\downann{0.28}} \\

\bluerow
Qwen3-Max
& \metric{0.83}{\downann{0.84}}
& \metric{0.96}{\downann{0.30}}
& \metric{1.50}{\sameann}
& \metric{7.14}{\downann{1.34}}
& \metric{20.62}{\upann{0.19}}
& \metric{9.49\textsuperscript{*}}{\sameann}
& \metric{16.46}{\upann{3.65}}
& \metric{7.42}{\downann{0.11}} \\

\bluerow
GPT-5.1-Codex-Max 
& \metric{1.67}{\sameann}
& \metric{1.07\textsuperscript{*}}{\downann{0.19}}
& \metric{1.50\textsuperscript{*}}{\sameann}
& \metric{15.35\textsuperscript{*}}{\upann{6.87}}
& \metric{20.00\textsuperscript{*}}{\downann{0.43}}
& \metric{6.14}{\downann{3.35}}
& \metric{5.79}{\downann{7.02}}
& \metric{7.65}{\upann{0.12}} \\

\bluerow
MiniMax-M2.1
& \metric{0.83}{\downann{0.84}}
& \metric{1.26\textsuperscript{*}}{\sameann}
& \metric{13.50}{\upann{12.00}}
& \metric{9.65}{\upann{1.17}}
& \metric{19.35\textsuperscript{$\dagger$}}{\downann{1.08}}
& \metric{9.49\textsuperscript{*}}{\sameann}
& \metric{21.65}{\upann{8.84}}
& \metric{9.33}{\upann{1.80}} \\

\bluerow
MiniMax-M2.5
& \metric{0.00}{\downann{1.67}}
& \metric{2.74\textsuperscript{*}}{\upann{1.48}}
& \metric{2.25}{\upann{0.75}}
& \metric{11.55}{\upann{3.07}}
& \metric{31.01\textsuperscript{*}}{\upann{10.58}}
& \metric{10.51}{\upann{1.02}}
& \metric{15.55}{\upann{2.74}}
& \metric{9.50}{\upann{1.97}} \\

\bluerow
Sonnet-4.5
& \metric{0.83}{\downann{0.84}}
& \metric{1.04}{\downann{0.22}}
& \metric{1.75}{\upann{0.25}}
& \metric{14.62}{\upann{6.14}}
& \metric{30.86}{\upann{10.43}}
& \metric{4.96}{\downann{4.53}}
& \metric{23.02}{\upann{10.21}}
& \metric{9.94}{\upann{2.41}} \\

\bluerow
GPT-5.4 (High)
& \metric{0.56}{\downann{1.11}}
& \metric{10.07}{\upann{8.81}}
& \metric{31.09}{\upann{29.59}}
& \metric{\best{27.98}}{\upann{19.50}}
& \metric{48.18}{\upann{27.75}}
& \metric{17.29}{\upann{7.80}}
& \metric{27.34}{\upann{14.53}}
& \metric{20.23}{\upann{12.70}} \\

\bluerow
GPT-5.2
& \metric{0.83}{\downann{0.84}}
& \metric{6.61}{\upann{5.35}}
& \metric{52.50}{\upann{51.00}}
& \metric{23.72}{\upann{15.24}}
& \metric{55.90}{\upann{35.47}}
& \metric{15.81}{\upann{6.32}}
& \metric{30.23}{\upann{17.42}}
& \metric{21.38}{\upann{13.85}} \\

\bluerow
Gemini-3.1-Pro
& \metric{3.89}{\upann{2.22}}
& \metric{7.42}{\upann{6.16}}
& \metric{62.84}{\upann{61.34}}
& \metric{18.53}{\upann{10.05}}
& \metric{45.51}{\upann{25.08}}
& \metric{14.48}{\upann{4.99}}
& \metric{40.19}{\upann{27.38}}
& \metric{21.59}{\upann{14.06}} \\

\bluerow
Opus-4.6
& \metric{\second{5.00}}{\upann{3.33}}
& \metric{7.78}{\upann{6.52}}
& \metric{75.92}{\upann{74.42}}
& \metric{25.52}{\upann{17.04}}
& \metric{41.04}{\upann{20.61}}
& \metric{\second{18.81}}{\upann{9.32}}
& \metric{24.75}{\upann{11.94}}
& \metric{23.16}{\upann{15.63}} \\

\bluerow
Opus-4.6 (1M)
& \metric{3.33}{\upann{1.66}}
& \metric{6.73}{\upann{5.47}}
& \metric{\second{77.16}}{\upann{75.66}}
& \metric{27.29}{\upann{18.81}}
& \metric{51.27}{\upann{30.84}}
& \metric{15.30}{\upann{5.81}}
& \metric{37.25}{\upann{24.44}}
& \metric{24.82}{\upann{17.29}} \\

\bluerow
Opus-4.7 (xHigh)
& \metric{\best{6.39}}{\upann{4.72}}
& \metric{\second{24.20}}{\upann{22.94}}
& \metric{76.75}{\upann{75.25}}
& \metric{\second{27.48}}{\upann{19.00}}
& \metric{\second{59.01}}{\upann{38.58}}
& \metric{16.53}{\upann{7.04}}
& \metric{\second{42.07}}{\upann{29.26}}
& \metric{\second{28.56}}{\upann{21.03}} \\

\midrule

\yellowgroupheader{9}{(IV) Evaluation with Proposed Method (Compared with Baseline)}

\lightyellowrowstart
\textbf{GLM-4.7 (OpenCode)}
& \metric{1.67}{\sameann}
& \metric{1.26}{\sameann}
& \metric{1.50}{\sameann}
& \metric{8.48}{\sameann}
& \metric{18.76}{\downann{1.67}}
& \metric{9.49}{\sameann}
& \metric{13.88}{\upann{1.07}}
& \metric{7.48}{\downann{0.05}} \\

\rowcolor{oursblue}
\textbf{GLM-4.7 \& ANDES (Ours)}
& \metric{3.90}{\upann{2.23}}
& \metric{\best{25.66}}{\upann{24.40}}
& \metric{\best{85.67}}{\upann{84.17}}
& \metric{27.12}{\upann{18.64}}
& \metric{\best{72.04}}{\upann{51.61}}
& \metric{\best{31.51}}{\upann{22.02}}
& \metric{\best{48.74}}{\upann{35.93}}
& \metric{\best{33.39}}{\upann{25.86}} \\

\bottomrule
\end{tabular}
\end{adjustbox}

\vspace{0.35em}
{\footnotesize
\textsuperscript{*} Model not submitted -- base model score shown
\hspace{1.5em}
\textsuperscript{$\dagger$} Evaluation error -- base model score shown
}
\end{table*}

\section{Experiments}
\label{sec:results}
To validate the effectiveness of \textsc{Andes}, we conduct systematic experiments across different base models and target benchmarks. First, we evaluate \textsc{Andes} under the official PostTrainBench setting, where it achieves state-of-the-art performance in autonomous post-training, demonstrating the overall effectiveness of our method. Subsequent component analyses and ablation studies decouple the sources of our performance gains. These results validate the efficacy of our interactive feedback loop and demonstrate the vast exploration capacity of the world tree routing mechanism. Finally, to demonstrate that \textsc{Andes} capability extends beyond single-target synthesis, we provide extensive evaluations on five additional benchmarks in Appendix~\ref{sec:appendix_multi_bench}. These results further verify the generality, efficiency, and cross-task generalization ability of our synthesized data. 

\subsection{Experimental Setup}
\label{subsec:exp_setup}

\paragraph{Trainer Agent Scaffold}
To ensure the proper execution of basic training procedures, we construct a lightweight, CLI-based trainer agent scaffold. We provide it with a fundamental scaffold necessary to execute training workflows, encompassing essential tools (e.g., \texttt{bash}, \texttt{read}, \texttt{write}), sub-agents such as a training monitor, and a skill system designed for the integration of \textsc{Andes}. To ensure fairness, we evaluated the performance of this scaffold independently in Section \ref{subsec:ablation}. We select GLM-4.7 as the driving model, as its moderate agentic capabilities provide an ideal comparative baseline to demonstrate our method's efficacy.

\paragraph{Evaluation}
Our experiments are primarily conducted on PostTrainBench\cite{posttrainbench_2026}. Following its official protocols, we deploy a sandbox environment on a single GPU with a 10-hour wall-clock time limit. We evaluate all four officially supported base models (Qwen3-1.7B, Qwen3-4B, SmolLM3-3B, and Gemma-3-4B) across seven benchmarks. These span mathematical reasoning (GSM8K~\cite{cobbe2021trainingverifierssolvemath}, AIME~2025), code generation (HumanEval~\cite{chen2021evaluatinglargelanguagemodels}), tool use (BFCL~\cite{patil2025bfcl}), scientific QA (GPQA-Main~\cite{rein2023gpqagraduatelevelgoogleproofqa}), medical dialogue (HealthBench~\cite{arora2025healthbenchevaluatinglargelanguage}), and open-ended writing (ArenaHardWriting~\cite{arenahard2024}). For every (model, benchmark) pair, we run the trainer agent end-to-end (28 runs in total) and report the official metric of the corresponding task. Full settings are detailed in Appendix \ref{details}.

\begin{figure}[t]
    \centering
    \includegraphics[
        width=\linewidth,
        height=0.19\textheight
    ]{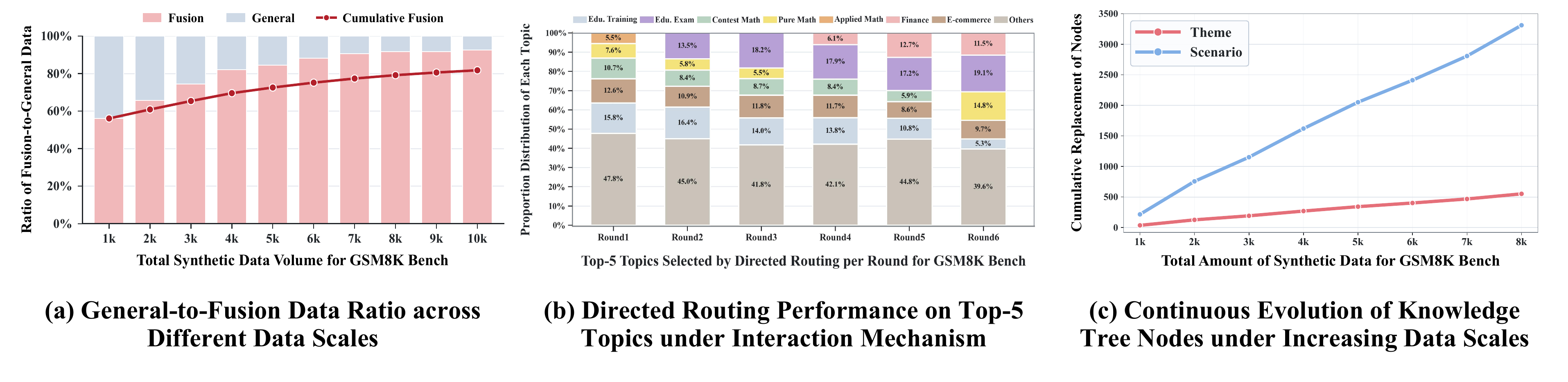}

    \vspace{-2mm}
    \caption{\textbf{Visualization of \textsc{Andes} routing and node evolution mechanism.}
    (a) The increasing fusion-data ratio shows that routing gradually shifts toward GSM8K-relevant nodes.(b) Top-routed GSM8K topics contain higher fusion ratios, indicating effective allocation to target-aligned capability regions.(c) The growing number of evolved themes and scenarios shows that \textsc{Andes} refreshes frequently selected nodes to preserve data diversity.}
    \label{fig:mechanism_visualization}
\end{figure}

\subsection{Performance in Autonomous Post-Training}

\begin{wrapfigure}[20]{r}{0.39\textwidth}
    \centering
    \vspace{-3mm}
    \includegraphics[width=0.88\linewidth]{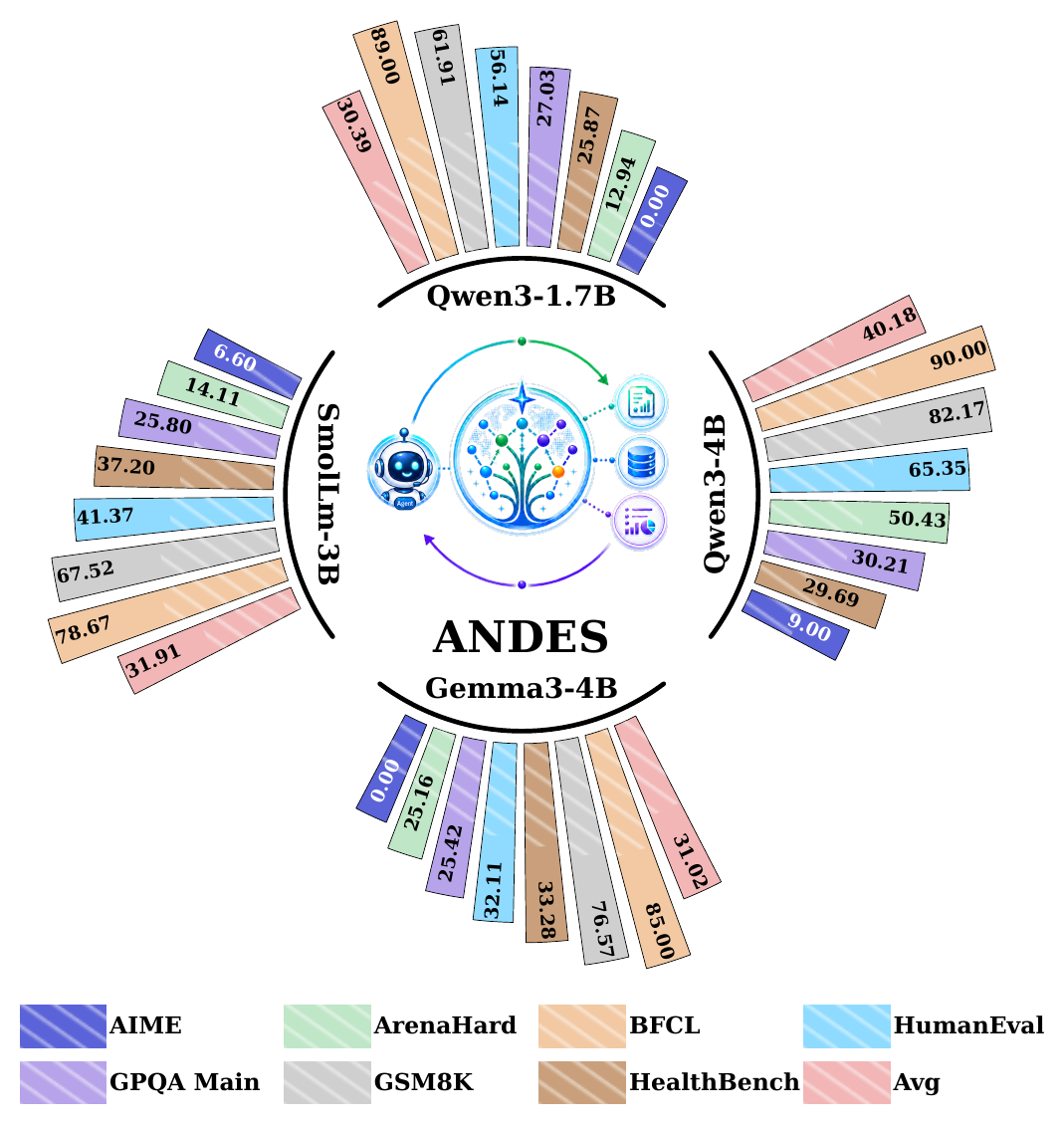}
    \vspace{-2mm}
    \caption{\textbf{Experimental results across four base models on PostTrainBench.}
    Different colors denote different benchmarks and the average score. \textsc{Andes} achieves the best average performance under three base-model settings.}
    \label{fig:four_experiment_result}
    \vspace{-5mm}
\end{wrapfigure}
\textbf{Autonomous Post-Training on PostTrainBench.} The results on PostTrainBench are reported in Tab.~\ref{tab:leaderboard_results} and Fig.~\ref{fig:four_experiment_result}. Across four base models and seven benchmarks, \textsc{Andes} achieves an average score of 33.39\%, outperforming the strongest competing method, Opus-4.7 (xHigh), which obtains 28.56\%, and thus establishes a new state-of-the-art result (detailed performance in Appendix \ref{four_sub_experiments}). Moreover, \textsc{Andes} consistently achieves the highest average performance under each base-model setting, indicating that its efficacy is not tied to a specific model architecture or parameter scale. Notably, when tasked with autonomous training, standard agents predominantly rely on conventional data acquisition methods (e.g., searching HuggingFace) and static data-preparation strategies, which inherently limits their potential. Specifically, the baseline GLM-4.7 (equipped with the default OpenCode CLI) severely struggles to navigate the complex, long-horizon execution required to complete the full training pipeline. Consequently, it yields an average score of only 7.48\%, remaining virtually stagnant compared to the 7.53\% zero-shot average of the unaligned base models. In stark contrast, after introducing our custom agent scaffold and the \textsc{Andes} tool, the average performance of GLM-4.7 surges significantly to 33.39\%, securing substantial gains across diverse domains such as BFCL, GSM8K and ArenaHard. These results demonstrate that \textsc{Andes} can effectively transform target capability diagnoses into high-quality, diverse, and trainable data, thereby significantly improving the effectiveness of autonomous post-training.

\subsection{Scalability Synthesis Performance}
\label{subsec:generalization}

\textbf{Analysis of Directed Routing and Node Evolution.} 
To understand how \textsc{Andes} simultaneously achieves precise benchmark alignment and structural diversity, we visualize its internal dynamics during a GSM8K synthesis session (Fig.~\ref{fig:mechanism_visualization}). Rather than uniformly sampling the World Tree, the directed routing mechanism actively shifts its focus according to the target task. As synthesis progresses, the system increasingly allocates its budget to target-relevant nodes, driving a steady rise in the proportion of ``fusion data'' while general data diminishes (Fig.~\ref{fig:mechanism_visualization}a). This indicates that routing becomes progressively more selective across synthesis rounds. A concrete breakdown of this behavior (Fig.~\ref{fig:mechanism_visualization}b) shows that \textsc{Andes} homes in on high-value regions, such as mathematical reasoning and quantitative relations. By constraining generation within these localized topics, the framework injects explicit reasoning logic while preserving diverse contextual backgrounds, combining task-specific supervision with broad scenario coverage.

Crucially, this concentrated routing does not lead to diversity collapse, a common pitfall when specific scenarios are over-sampled. The node evolution mechanism counteracts this risk by continuously refreshing high-relevance regions of the World Tree. As the GSM8K-oriented data volume scales, the cumulative number of newly evolved themes and scenarios keeps increasing rather than hitting a ceiling (Fig.~\ref{fig:mechanism_visualization}c). Whenever highly relevant nodes are repeatedly queried, \textsc{Andes} spawns novel sub-scenarios instead of recycling exhausted templates, maintaining fresh yet task-relevant training contexts across synthesis rounds. Together, directed routing ensures cognitive alignment with the target task, while node evolution provides a continual supply of contextual diversity. As a result, \textsc{Andes} can continuously generate vertical training data without sacrificing diversity, supporting robust and scalable autonomous alignment.

\subsection{Ablation Study}
\label{subsec:ablation}

\begin{wrapfigure}[18]{r}{0.42\textwidth}
    \centering
    \vspace{-3mm}
    \includegraphics[width=0.90\linewidth]{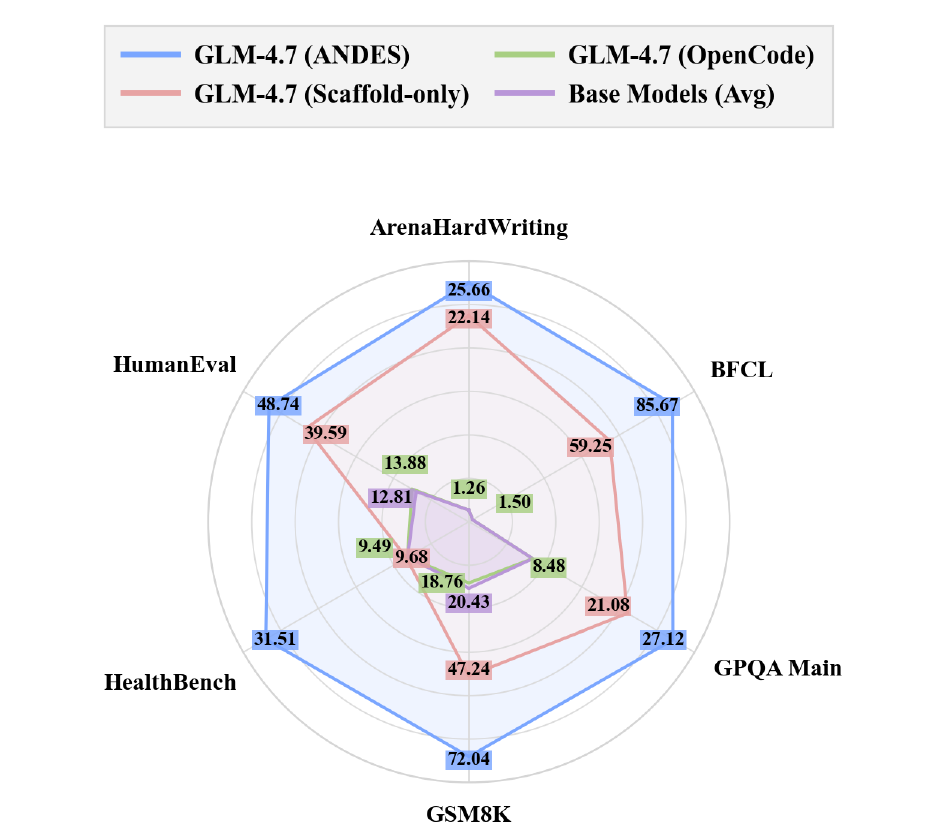}
    \vspace{-2mm}
    \caption{\textbf{Decoupled performance on PostTrainBench.} Compared to GLM-4.7 (Scaffold-only) (averaged 21.56\%), \textsc{Andes} drives an 11.83\% gain to 33.39\%, achieving multi-dimensional breakthroughs.}
    \label{fig:scaffold_ablation}
    \vspace{-5mm}
\end{wrapfigure}
\paragraph{Decoupling the Source of Improvements.} 
Fig.~\ref{fig:scaffold_ablation} decouples the performance gains achieved by our proposed framework (detailed performance in Appendix \ref{four_sub_experiments}). We evaluate four distinct settings across the four base models (Qwen3-1.7B, Qwen3-4B, SmolLM3-3B, and Gemma-3-4B) seperately: the original base model, a standard GLM-4.7 baseline (equipped with OpenCode CLI), GLM-4.7 equipped with our custom agent scaffold, and finally, equipped with the \textsc{Andes} tool. 

As shown, replacing the generic OpenCode baseline with our custom agent scaffold yields substantial initial gains—improving scores on BFCL (1.50 to 59.25), GSM8K (20.43 to 47.24), and HumanEval (12.81 to 39.59). This proves that specialized infrastructure is crucial to stabilize the autonomous training pipeline, handling data formatting and execution far better than standard environments. However, agents equipped with scaffolds remain limited to retrieving and curating similar data from web-based sources like Hugging Face; their data organization capabilities continue to pose a critical bottleneck that constrains further model performance gains. Integrating the \textsc{Andes} synthesis tool drives a definitive second wave of improvements across all benchmarks, with notable surges on BFCL (59.25 to 85.67), GSM8K (47.24 to 72.04) and HealthBench (9.68 to 31.51). This two-stage progression confirms our hypothesis: while the scaffold successfully operationalizes the physical workflow, it is the active, target-aligned data synthesis of \textsc{Andes} that ultimately pushes the model to state-of-the-art performance.

\textbf{Ablation on Report-driven Interaction Mechanism.}
Tab.~\ref{fig:data_efficiency} shows the ablation results of the report-driven interaction mechanism. 
After removing the interaction mechanism, the model still achieves 26.80\% average performance, which is much higher than the GLM-4.7 baseline of 7.11\%. 
This indicates that the basic synthesis ability of \textsc{Andes} already provides effective training signals. 
However, the full \textsc{Andes} system further improves the average score to 30.39\% and achieves better results on tasks such as BFCL and HumanEval. For example, the score increases from 68.00 to 89.00 on BFCL, from 50.01 to 56.14 on HumanEval.

These results show that one-shot data synthesis is insufficient to fully exploit the capability of \textsc{Andes}. 
With report-driven interaction, the trainer agent can use the returned synthesis report, including data quality feedback, effective sample count, and logical diversity signals, to adjust subsequent synthesis configurations. 
It can steer data generation toward under-covered capability regions. 
Therefore, the interaction mechanism improves both the quality and diversity of synthesized data, validating the effectiveness of our multi-round agent--tool interaction paradigm.

\begin{table*}[t]
\centering
\setlength{\tabcolsep}{3.9pt}
\renewcommand{\arraystretch}{1.18}
\caption{Experiments results on PostTrainBench (accuracy in \%) based on the Qwen3-1.7B base model. The best and second-best results among agent-based methods are highlighted in \textbf{bold} and \underline{underlined}. Numbers after arrows denote absolute changes over the corresponding zero-shot base model, with improvements in \textcolor{myred}{red} and declines in \textcolor{mygreen}{green}.}
\vspace{4pt}
\label{fig:data_efficiency}

\begin{adjustbox}{width=\textwidth,center}
\fontsize{12.3}{14.2}\selectfont
\begin{tabular}{l|cccccccc}
\toprule
\addlinespace[0.12em]
\textbf{Method} & \textbf{AIME 2025} & \textbf{ArenaHard} & \textbf{BFCL} & \textbf{GPQA Main} & \textbf{GSM8K} & \textbf{HealthBench} & \textbf{HumanEval} & \textbf{AVG (\%)} \\[0.16em]
\midrule

\groupheader{9}{(I) Evaluation with Base Models (Zero-Shot)}
\grouprow
\textbf{Qwen3-1.7B(Base)}
& 0.00 & 0.91 & 0.00 & 14.06 & 12.66 & 7.50 & 7.93 & 6.66 \\
\midrule

\bluegroupheader{9}{(II) Evaluation with Agent-Post-Trained Base Models)}

GPT-5.5 (Reprompted)
& \metric{0.00}{\sameann}
& \metric{0.91}{\sameann}
& \metric{100.00}{\upann{100.00}}
& \metric{29.46}{\upann{15.40}}
& \metric{69.90}{\upann{57.24}}
& \metric{0.00}{\downann{7.50}}
& \metric{60.98}{\upann{53.05}}
& \metric{27.18}{\upann{20.52}} \\

Opus-4.7
& \metric{6.67}{\upann{6.67}}
& \metric{33.84}{\upann{32.93}}
& \metric{91.33}{\upann{91.33}}
& \metric{29.32}{\upann{15.26}}
& \metric{50.47}{\upann{37.81}}
& \metric{16.23}{\upann{8.73}}
& \metric{42.28}{\upann{34.35}}
& \metric{30.17}{\upann{23.51}} \\

\midrule

\yellowgroupheader{9}{(IV) Evaluation with Proposed Method (Compared with Baseline)}

\baselinebluerowstart
\textbf{GLM-4.7 (OpenCode)}
& \metric{0.00}{\sameann}
& \metric{0.91}{\sameann}
& \metric{0.00}{\sameann}
& \metric{14.06}{\sameann}
& \metric{12.66}{\sameann}
& \metric{7.54}{\upann{0.04}}
& \metric{12.20}{\upann{4.27}}
& \metric{7.11}{\upann{0.45}} \\

\withoutintbluerowstart
\textbf{Ours without Interaction}
& \metric{0.00}{\sameann}
& \metric{\second{8.02}}{\upann{7.11}}
& \metric{\second{68.00}}{\upann{68.00}}
& \metric{25.41}{\upann{11.35}}
& \metric{\second{59.86}}{\upann{47.20}}
& \metric{\second{22.61}}{\upann{15.11}}
& \metric{\second{50.01}}{\upann{42.08}}
& \metric{\second{26.80}}{\upann{20.14}} \\

\oursbluerowstart
\textbf{GLM-4.7 \& ANDES (Ours)}
& \metric{0.00}{\sameann}
& \metric{\best{12.94}}{\upann{12.03}}
& \metric{\best{89.00}}{\upann{89.00}}
& \metric{27.03}{\upann{12.97}}
& \metric{\best{61.91}}{\upann{49.25}}
& \metric{\best{25.87}}{\upann{18.37}}
& \metric{\best{56.14}}{\upann{48.21}}
& \metric{\best{30.39}}{\upann{23.73}} \\
\bottomrule
\end{tabular}
\end{adjustbox}

\end{table*}

\subsection{Diversity and Feature Space Analysis}
\label{subsec:diversity_tsne}
\textbf{Breadth and Universality of The World Tree.}
Beyond directed routing and node evolution, the effectiveness of \textsc{Andes} also relies on the broad synthesis space provided by its World Tree. The World Tree covers diverse macro-level knowledge domains and organizes them into a hierarchical topic-theme-scenario structure, containing 72 topics, 394 themes, and 1,182 scenarios. This large-scale hierarchy prevents data synthesis from being restricted to a single task domain or a fixed set of templates. As a result, \textsc{Andes} can preserve general contextual coverage while routing synthesis toward task-relevant nodes for vertical capability improvement. The feature-space visualization further shows that synthetic samples for different benchmarks form distinguishable task-related regions while still sharing overlapping areas, indicating that the World Tree supports both target alignment and cross-task generality. We provide a more detailed analysis of the World Tree coverage, hierarchical composition, and feature-space distribution in Appendix~\ref{world_tree}.

\section{Conclusion}
\label{sec:Conclusion}
In this work, we introduced \textsc{Andes}, an agent-native data evolving synthesis tool for autonomous instruction alignment.
We argue that the key bottleneck in automated post-training lies not only in executing training, but in reliably acquiring high-quality, targeted, and diverse data under limited compute budgets.
\textsc{Andes} addresses this by reformulating data synthesis as a plug-and-play agent skill, combining target-driven requests, self-evolving World Tree routing, two-stage generation and refinement, and report-driven closed-loop interaction to continuously steer synthesis toward required capabilities while preserving contextual diversity.
Experiments on PostTrainBench show that even a foundationally weaker trainer agent equipped with \textsc{Andes} achieves state-of-the-art post-training performance across multiple base models, with ablation studies validating the contribution of each component.
Overall, \textsc{Andes} demonstrates that structured, interactive, and evolvable data synthesis interfaces are a promising step toward scalable and reliable automated model alignment.

\clearpage
\bibliographystyle{plain}
\bibliography{reference}

\appendix
\appendix
\clearpage 

\section*{Appendix Content}
\startcontents[sections]
\printcontents[sections]{l}{1}{\setcounter{tocdepth}{2}}
\vspace{2em} 

\section{Implementation Details}
\label{details}

We provide the detailed experimental settings in this appendix. We organize them into four parts: the sandbox in which the trainer agent runs, the agent scaffold itself, ANDES internals, and the training/evaluation pipeline.

\subsection{Sandbox and Hardware}
\label{app:sandbox}

To ensure rigorous and reproducible evaluations, all PostTrainBench experiments are conducted within an isolated, standardized sandbox environment. We abstract away low-level dependency management by exposing unified interfaces to the trainer agent, allowing it to focus entirely on optimization strategies rather than system-level troubleshooting. Each post-training session is strictly constrained to a 10-hour compute budget on a single A100 GPU, enforcing realistic resource bounds for autonomous alignment. Upon timeout, only the designated final checkpoint is extracted for assessment. To strictly prevent train-test leakage or environmental manipulation, downstream evaluations are executed in an independent, pristine sandbox. Furthermore, in accordance with the PostTrainBench protocol, an independent auditor module (powered by GPT-5.1) monitors the agent's trajectory to detect and prevent reward-hacking behaviors, such as evaluation tampering or unauthorized model substitution, thereby guaranteeing the integrity of the final results.

\subsection{Agent Scaffold }
\label{app:scaffold}

\paragraph{Tool action space.} The trainer agent operates over a fixed set of primitives, summarized in Table~\ref{tab:tools}. \texttt{bash} runs in two modes: foreground commands time out after 1\,800~s with a hint to retry in background mode, while background commands are launched via \texttt{nohup}, registered in an internal job manager, and observed through the \texttt{wait} and \texttt{read\_file} tools. \texttt{search\_dataset} wraps the HuggingFace Hub API and is reserved for the \emph{data\_loader} sub-agent.

\begin{table}[h]
\centering
\caption{Tool action space exposed to the trainer agent.}
\label{tab:tools}
\small
\begin{tabular}{ll}
\toprule
Tool & Purpose \\
\midrule
\texttt{bash} & Shell execution (foreground / background modes) \\
\texttt{read\_file} & Range or tail reads on workspace files (15\,KB output cap) \\
\texttt{write\_file} / \texttt{edit} & Workspace I/O and exact-match edits \\
\texttt{search\_dataset} & HuggingFace Hub dataset retrieval \\
\texttt{wait} & Synchronization for long-running jobs (capped at 3\,600~s) \\
\texttt{todo} & Multi-step plan tracking \\
\texttt{task} & Dispatch a sub-agent (\texttt{data\_loader} or \texttt{training\_monitor}) \\
\texttt{Skill} & Load a skill file into the conversation \\
\bottomrule
\end{tabular}
\end{table}

\paragraph{Sub-agents.} Two sub-agents are pre-registered. The \emph{data\_loader} (tools: \texttt{search\_dataset}, \texttt{read\_file}, \texttt{write\_file}, \texttt{bash}, \texttt{Skill}) is responsible for dataset discovery, sampling-based audit and Alpaca-format normalization, and is the entry point through which ANDES is invoked. The \emph{training\_monitor} (tools: \texttt{bash}, \texttt{read\_file}, \texttt{wait}, \texttt{Skill}) supervises a launched training job, parses ETA from the training log, and verifies that artifacts exist before reporting back. Both sub-agents communicate with the trainer agent only via a file-based prompt, ensuring a clean separation of context.

\paragraph{Skills.} Capability-specific guidance is delivered through three Markdown skill files loaded on demand: \texttt{posttrain skill} (engineering principles for the full training lifecycle), \texttt{llamafactory skill} (dataset registration, YAML template, training settings), and \texttt{data-strategy skill} (the ANDES skill: domain decomposition, report-driven discard ratio decisions, format-protocol selection rules, and seven failure-signal taxonomy). Note that we provide only basic llamafactory configuration options to ensure that training proceeds normally, thereby isolating the contribution of \textsc{Andes}). The ANDES skill enforces (i) macro-level domain decomposition into $\sim$6 cognitive-operation domains, (ii) one ANDES call per domain with 1000--2000 raw samples and discard ratio $\in \{0\%, 10\%, 20\%\}$ chosen per round from the report.

\subsection{ANDES Internals}
\label{app:andes_internals}
We build our code upon DataFlow\cite{liang2025dataflowllmdrivenframeworkunified} framework.

\paragraph{Internal LLMs.} The router, generator, refiner, evolver, and diversity summarizer are all instantiated with gpt-4o at temperature $0$, sharing a single API gateway.

\paragraph{World tree.} The initial taxonomy contains $72$ topics, $394$ themes, and $1\,182$ domains, grouped into $8$ macro categories (Healthcare and Life Sciences; Mathematics and Natural Sciences; Computing, AI and Information Security; Engineering, Technology and Hardware Systems; Education, Family and Personal Development; Business, Governance and Public Affairs; Humanities, Arts and Entertainment; Lifestyle, Consumption and Leisure Services).

\paragraph{Routing parameters.} Topics are weighted-sampled in mini-batches of $B=5$, with all topic weights initialized to $1.0$. After each batch is rated by the router, the reward factor for \emph{Strong}-rated topics is $\gamma^{+}=1.5$ and the decay factor for \emph{Weak}-rated topics is $\gamma^{-}=0.8$, clipped at a floor $\epsilon=0.1$. \emph{Ambiguous} samples are assigned to the fusion track with probability $0.8$. Each ANDES call routes $\lceil N/3 \rceil$ scenarios for $N$ requested samples.

\paragraph{Subtree evolution.} A topic triggers expansion once its cumulative selection count exceeds $\rho=0.8$ times the current size of its theme pool. The evolver is prompted with the topic, the task description, and the full history of previously generated subtrees, and is required to return $6$ new sub-domains, each with $6$ themes (i.e., $36$ new scenarios per evolution). After expansion the topic's selection count and weight are reset to their initial values to prevent cascaded re-evolution.

\paragraph{Refinement.} Stage~2 first asks the refiner to critique each QA pair and emit an effort score $e_i \in \{1,\dots,5\}$. Samples with $e_i > 4$ are dropped before any rewriting; the rest are rewritten conditioned on the original pair and the critique. The diversity check then samples mini-batches of $25$ items from the fusion-track subset, targeting $\sim$$20\%$ coverage of the fusion pool, and aggregates per-batch \textit{Low/Medium/High} verdicts plus a free-form pattern summary into the report returned to the trainer agent.

\paragraph{Format protocols.} The trainer agent selects \texttt{format\_requirement} once per benchmark from \{\texttt{unstructured}, \texttt{code}, \texttt{tool\_call}\}; the protocol is attached only to the answer prompt of fusion-track samples and constrains the output to contain at least one fenced code block of the corresponding type. \texttt{format\_requirement} is fixed across rounds within the same benchmark.

\section{Extended Evaluations on Diverse Downstream Tasks}
\label{sec:appendix_multi_bench}

While the main text demonstrates the efficacy of \textsc{Andes} in the standard autonomous PostTrainBench setting (single benchmark per training), a truly robust data synthesis framework must be capable of generalizing across diverse tasks without suffering from capability degradation. In this section, we extend our evaluation to validate the cross-task generalization and multi-target synthesis capabilities of \textsc{Andes}. Specifically, we prompt the trainer agent to utilize \textsc{Andes} to simultaneously acquire a balanced dataset that covers all target benchmarks.

\subsection{Experimental Setup}
We employ the trainer agent to simultaneously synthesize data targeting five benchmarks (AIME24, GaoKao\cite{zhong2023agieval}, MBPP\cite{austin2021program}, MMLU\cite{hendryckstest2021}, CEVAL\cite{huang2023ceval}) in a single session. We then evaluate and report the SFT performance on the Qwen3-8B model, comparing our results against other high-quality dataset baselines: 
(1) Infininstruct\cite{li2025infinityinstructscalinginstruction}, a large-scale dataset containing millions of high-quality instruction-tuning data.
(2) Dataflow\cite{liang2025dataflowllmdrivenframeworkunified}, a static two-stage generic data synthesis pipeline.
We synthesis 10,000 data using GLM-4.7+\textsc{Andes}, and sampled 10,000 training data points from the baseline datasets, trained for one epoch with learning rate 2e-5. To further compare data efficiency, we also compared the training results using 1 million data from the Infininstruct dataset.

\subsection{Results and Discussion}
\begin{table*}[t]
\centering
\setlength{\tabcolsep}{5.0pt}
\renewcommand{\arraystretch}{1.16}
\caption{Results across five diverse benchmarks. The best and second-best results are highlighted in \textbf{bold} and \underline{underlined}, respectively. Numbers after arrows denote absolute changes over the corresponding base model with improvements in \textcolor{myred}{red} and declines in \textcolor{mygreen}{green}.}
\label{tab:reasoning_results}

\begin{adjustbox}{max width=\textwidth,center}
\fontsize{12.3}{14.2}\selectfont
\begin{tabular}{lc|ccccc|c}
\toprule
\addlinespace[0.12em]
\textbf{Method}
& \textbf{Data Volume}
& \textbf{AIME24}
& \textbf{Gaokao}
& \textbf{MBPP}
& \textbf{MMLU}
& \textbf{CEval}
& \textbf{Overall}
\\[0.16em]
\midrule

\rowcolor{sectiongray}

Base Model(Qwen3-8B-Base)
& --
& 13.3
& 35.2
& 82.5
& 73.0
& 82.5
& 57.3
\\

\hspace{1em}+ Infinstruct
& 10k
& \metric{0.0}{\downann{13.3}}
& \metric{\second{34.1}}{\downann{1.1}}
& \metric{76.2}{\downann{6.3}}
& \metric{74.2}{\upann{1.2}}
& \metric{81.3}{\downann{1.2}}
& \metric{53.2}{\downann{4.1}}
\\

\hspace{1em}+ Infinstruct
& 1M
& \metric{0.0}{\downann{13.3}}
& \metric{28.6}{\downann{6.6}}
& \metric{70.1}{\downann{12.4}}
& \metric{71.3}{\downann{1.7}}
& \metric{77.9}{\downann{4.6}}
& \metric{49.6}{\downann{7.7}}
\\

\hspace{1em}+ Dataflow
& 10k
& \metric{\best{10.0}}{\downann{3.3}}
& \metric{31.9}{\downann{3.3}}
& \metric{\second{77.2}}{\downann{5.3}}
& \metric{\second{74.6}}{\upann{1.6}}
& \metric{\second{82.1}}{\downann{0.4}}
& \metric{\second{55.2}}{\downann{2.1}}
\\

\rowcolor{oursblue}
\hspace{1em}+ \textbf{ANDES (Ours)}
& 10k
& \metric{\best{10.0}}{\downann{3.3}}
& \metric{\best{49.5}}{\upann{14.3}}
& \metric{\best{77.7}}{\downann{4.8}}
& \metric{\best{74.8}}{\upann{1.8}}
& \metric{\best{82.4}}{\downann{0.1}}
& \metric{\best{58.9}}{\upann{1.6}}
\\

\bottomrule
\end{tabular}
\end{adjustbox}
\end{table*}
As shown in Table~\ref{tab:reasoning_results}, \textsc{Andes} significantly outperforms both the 10K-sampled baselines and the 1M full-scale static dataset across the diverse evaluation suite. Notably, conventional static pipelines and large-scale data mixtures (such as Infininstruct and Dataflow) often struggle with capability degradation during multi-target SFT, leading to noticeable performance drops compared to the base model. In contrast, \textsc{Andes} not only effectively mitigates these severe regressions but also drives exceptional gains in specific domains. Particularly striking is its performance on the Gaokao benchmark, achieving a substantial absolute improvement of 14.3 points while all other static baselines suffered severe degradation. This remarkable contrast underscores the critical necessity of actively acquiring targeted, in-domain data, demonstrating that \textsc{Andes}'s dynamic routing and scenario abstraction can seamlessly adapt to localized knowledge requirements to yield a comprehensively balanced curriculum for robust cross-task generalization.

\begin{table*}[t]
    \centering
    \setlength{\tabcolsep}{3.9pt}
    \renewcommand{\arraystretch}{1.18}
    \caption{Experiments results on PostTrainBench (accuracy in \%) based on the Qwen3-1.7B base model. The best and second-best results are highlighted in \textbf{bold} and \underline{underlined}. Numbers after arrows denote absolute changes over the corresponding zero-shot base model, with improvements in \textcolor{myred}{red} and declines in \textcolor{mygreen}{green}.}
    \vspace{4pt}
    \label{tab:qwen3-1.7b}
    
    \begin{adjustbox}{width=\textwidth,center}
    \fontsize{12.3}{14.2}\selectfont
    \begin{tabular}{l|cccccccc}
    \toprule
    \addlinespace[0.12em]
    \textbf{Method} & \textbf{AIME 2025} & \textbf{ArenaHard} & \textbf{BFCL} & \textbf{GPQA Main} & \textbf{GSM8K} & \textbf{HealthBench} & \textbf{HumanEval} & \textbf{AVG (\%)} \\[0.16em]
    \midrule
    
    \groupheader{9}{(I) Evaluation with Official Instruct Model}
    \grouprow
    \textbf{Official Instruct Model}
    & 26.67 & 50.00 & 94.00 & 35.49 & 88.48 & 44.92 & 68.90 & 49.41 \\
    \midrule
    
    \groupheader{9}{(II) Evaluation with Base Model (Zero-Shot)}
    \grouprow
    \textbf{Base Model (Qwen3-1.7B)}
    & 0.00 & 0.91 & 0.00 & 14.06 & 12.66 & 7.54 & 7.93 & 6.66 \\
    \midrule
    
    \bluegroupheader{9}{(III) Evaluation with Agent-Post-Trained Base Model}
    
    \bluerow
    Sonnet-4.5
    & \metric{0.00}{\sameann}
    & \metric{0.21}{\downann{0.70}}
    & \metric{0.00}{\sameann}
    & \metric{16.74}{\upann{2.68}}
    & \metric{2.58}{\downann{10.08}}
    & \metric{0.00}{\downann{7.54}}
    & \metric{0.61}{\downann{7.32}}
    & \metric{4.09}{\downann{2.57}} \\
    
    \bluerow
    MiniMax-M2.5
    & \metric{0.00}{\sameann}
    & \metric{1.76\textsuperscript{*}}{\upann{0.85}}
    & \metric{0.00}{\sameann}
    & \metric{14.06}{\sameann}
    & \metric{12.66\textsuperscript{*}}{\sameann}
    & \metric{7.66}{\upann{0.12}}
    & \metric{7.93}{\sameann}
    & \metric{6.75}{\upann{0.09}} \\
    
    \bluerow
    MiniMax-M2.1
    & \metric{0.00}{\sameann}
    & \metric{0.91\textsuperscript{*}}{\sameann}
    & \metric{0.00}{\sameann}
    & \metric{14.51}{\upann{0.45}}
    & \metric{11.52\textsuperscript{$\dagger$}}{\downann{1.14}}
    & \metric{7.54\textsuperscript{*}}{\sameann}
    & \metric{12.20}{\upann{4.27}}
    & \metric{7.10}{\upann{0.44}} \\
    
    \bluerow
    Qwen3-Max
    & \metric{0.00}{\sameann}
    & \metric{0.91}{\sameann}
    & \metric{0.00}{\sameann}
    & \metric{14.06}{\sameann}
    & \metric{12.66}{\sameann}
    & \metric{7.54\textsuperscript{*}}{\sameann}
    & \metric{12.80}{\upann{4.87}}
    & \metric{7.17}{\upann{0.51}} \\
    
    \bluerow
    Kimi-K2-Thinking
    & \metric{0.00\textsuperscript{*}}{\sameann}
    & \metric{0.91\textsuperscript{*}}{\sameann}
    & \metric{0.00\textsuperscript{*}}{\sameann}
    & \metric{14.06\textsuperscript{*}}{\sameann}
    & \metric{12.66}{\sameann}
    & \metric{7.54\textsuperscript{$\dagger$}}{\sameann}
    & \metric{17.07\textsuperscript{*}}{\upann{9.14}}
    & \metric{7.63}{\upann{0.97}} \\
    
    \bluerow
    GPT-5.1-Codex-Max 
    & \metric{0.00}{\sameann}
    & \metric{0.14\textsuperscript{*}}{\downann{0.77}}
    & \metric{0.00\textsuperscript{*}}{\sameann}
    & \metric{24.78\textsuperscript{*}}{\upann{10.72}}
    & \metric{10.92\textsuperscript{*}}{\downann{1.74}}
    & \metric{7.54}{\sameann}
    & \metric{6.10}{\downann{1.83}}
    & \metric{8.64}{\upann{1.98}} \\
    
    \bluerow
    Opus-4.6
    & \metric{0.00}{\sameann}
    & \metric{1.15}{\upann{0.24}}
    & \metric{28.33}{\upann{28.33}}
    & \metric{22.92}{\upann{8.86}}
    & \metric{27.14}{\upann{14.48}}
    & \metric{8.58}{\upann{1.04}}
    & \metric{22.76}{\upann{14.83}}
    & \metric{13.90}{\upann{7.24}} \\
    
    \bluerow
    GPT-5.2
    & \metric{2.22}{\upann{2.22}}
    & \metric{1.27}{\upann{0.36}}
    & \metric{29.33}{\upann{29.33}}
    & \metric{17.41}{\upann{3.35}}
    & \metric{51.00}{\upann{38.34}}
    & \metric{9.33}{\upann{1.79}}
    & \metric{32.72}{\upann{24.79}}
    & \metric{16.68}{\upann{10.02}} \\
    
    \bluerow
    Gemini-3.1-Pro
    & \metric{0.00}{\sameann}
    & \metric{2.27}{\upann{1.36}}
    & \metric{85.67}{\upann{85.67}}
    & \metric{16.15}{\upann{2.09}}
    & \metric{39.54}{\upann{26.88}}
    & \metric{10.93}{\upann{3.39}}
    & \metric{36.18}{\upann{28.25}}
    & \metric{19.78}{\upann{13.12}} \\
    
    \bluerow
    GPT-5.4 (High)
    & \metric{3.33}{\upann{3.33}}
    & \metric{0.91}{\sameann}
    & \metric{0.00}{\sameann}
    & \metric{24.33}{\upann{10.27}}
    & \metric{\best{73.92}}{\upann{61.26}}
    & \metric{\best{29.39}}{\upann{21.85}}
    & \metric{33.54}{\upann{25.61}}
    & \metric{22.19}{\upann{15.53}} \\
    
    \bluerow
    Opus-4.6 (1M)
    & \metric{\second{5.56}}{\upann{5.56}}
    & \metric{3.21}{\upann{2.30}}
    & \metric{87.33}{\upann{87.33}}
    & \metric{22.99}{\upann{8.93}}
    & \metric{42.99}{\upann{30.33}}
    & \metric{9.73}{\upann{2.19}}
    & \metric{37.20}{\upann{29.27}}
    & \metric{22.99}{\upann{16.33}} \\
    
    \bluerow
    Opus-4.7 (xHigh)
    & \metric{\best{6.67}}{\upann{6.67}}
    & \metric{\best{33.84}}{\upann{32.93}}
    & \metric{\best{91.33}}{\upann{91.33}}
    & \metric{\best{29.32}}{\upann{15.26}}
    & \metric{50.47}{\upann{37.81}}
    & \metric{16.23}{\upann{8.69}}
    & \metric{\second{42.28}}{\upann{34.35}}
    & \metric{\second{30.17}}{\upann{23.51}} \\
    
    \midrule
    
    \yellowgroupheader{9}{(IV) Evaluation with Proposed Method (Compared with Baseline)}
    
    \lightyellowrowstart
    \textbf{GLM-4.7 (OpenCode)}
    & \metric{0.00\textsuperscript{*}}{\sameann}
    & \metric{0.91\textsuperscript{*}}{\sameann}
    & \metric{0.00\textsuperscript{*}}{\sameann}
    & \metric{14.06\textsuperscript{$\dagger$}}{\sameann}
    & \metric{12.66\textsuperscript{*}}{\sameann}
    & \metric{7.54\textsuperscript{*}}{\sameann}
    & \metric{12.20}{\upann{4.27}}
    & \metric{7.11}{\upann{0.45}} \\

    \lightyellowrowstart
    \textbf{GLM-4.7 (Scaffold-only)}
    & \metric{0.00}{\sameann}
    & \metric{5.70}{\upann{4.79}}
    & \metric{43.00}{\upann{43.00}}
    & \metric{20.13}{\upann{6.07}}
    & \metric{36.88}{\upann{24.22}}
    & \metric{16.74}{\upann{9.20}}
    & \metric{44.10}{\upann{36.17}}
    & \metric{19.45}{\upann{12.79}} \\

    \rowcolor{oursblue}
    \textbf{GLM-4.7 \& ANDES (Ours)}
    & \metric{0.00}{\sameann}
    & \metric{\second{12.94}}{\upann{12.03}}
    & \metric{\second{89.00}}{\upann{89.00}}
    & \metric{\second{27.03}}{\upann{12.97}}
    & \metric{\second{61.91}}{\upann{49.25}}
    & \metric{\second{25.87}}{\upann{18.33}}
    & \metric{\best{56.14}}{\upann{48.21}}
    & \metric{\best{30.39}}{\upann{23.73}} \\
    
    \bottomrule
    \end{tabular}
    \end{adjustbox}
    
    \vspace{0.35em}
    {\footnotesize
    \textsuperscript{*} Model not submitted -- base model score shown
    \hspace{1.5em}
    \textsuperscript{$\dagger$} Evaluation error -- base model score shown
    }
    \end{table*}

\section{Four sub-experiments on PostTrainBench}
\label{four_sub_experiments}
\textbf{Sub-experiments on Different Base Models.}Tabs.~\ref{tab:qwen3-1.7b}, \ref{tab:qwen3-4b}, \ref{tab:smollm3-3b}, and \ref{tab:gemma3-4b} report the detailed PostTrainBench results under four base-model settings. 
Overall, \textsc{Andes} brings consistent improvements across different model families and different initial capability levels. 
For Qwen3-1.7B, \textsc{Andes} improves the average score from 7.11\% of the GLM-4.7 baseline to 30.39\%, surpassing Opus-4.7 with 30.17\% and achieving the best result. 
For SmolLM3-3B and Gemma3-4B, \textsc{Andes} achieves average scores of 31.91\% and 31.02\%, respectively, outperforming all agent-based comparison methods. 
For Qwen3-4B, \textsc{Andes} obtains 40.18\%, which is slightly lower than GPT-5.4 (High) with 41.40\%, but still substantially improves over the GLM-4.7 baseline of 14.70\% and achieves strong competitive performance on multiple benchmarks.

These results show that the effectiveness of \textsc{Andes} does not depend on a specific base model. 
Across Qwen, SmolLM, and Gemma model families, \textsc{Andes} can synthesize useful training data according to the target benchmark requirements. 
Moreover, the improvements cover diverse tasks, including mathematical reasoning, tool use, code generation, health dialogue, and open-ended writing. 
This indicates that \textsc{Andes} does not merely optimize for a single benchmark, but provides general and transferable training signals across different tasks. Furthermore, the metrics for the "scaffold-only" agent across four models and seven benchmarks were consistently lower than those of ANDES. This further substantiates the net gains that ANDES contributes to the trainer agent.

\begin{table*}[t]
    \centering
    \setlength{\tabcolsep}{3.9pt}
    \renewcommand{\arraystretch}{1.18}
    \caption{Experiments results on PostTrainBench (accuracy in \%) based on the Qwen3-4B base model. The best and second-best results are highlighted in \textbf{bold} and \underline{underlined}. Numbers after arrows denote absolute changes over the corresponding zero-shot base model, with improvements in \textcolor{myred}{red} and declines in \textcolor{mygreen}{green}.}
    \vspace{4pt}
    \label{tab:qwen3-4b}
    
    \begin{adjustbox}{width=\textwidth,center}
    \fontsize{12.3}{14.2}\selectfont
    \begin{tabular}{l|cccccccc}
    \toprule
    \addlinespace[0.12em]
    \textbf{Method} & \textbf{AIME 2025} & \textbf{ArenaHard} & \textbf{BFCL} & \textbf{GPQA Main} & \textbf{GSM8K} & \textbf{HealthBench} & \textbf{HumanEval} & \textbf{AVG (\%)} \\[0.16em]
    \midrule
    
    \groupheader{9}{(I) Evaluation with Official Instruct Model}
    \grouprow
    \textbf{Official Instruct Model}
    & 53.33 & 86.84 & 95.00 & 44.64 & 93.78 & 52.72 & 77.44 & 63.75 \\
    \midrule
    
    \groupheader{9}{(II) Evaluation with Base Model (Zero-Shot)}
    \grouprow
    \textbf{Base Model (Qwen3-4B)}
    & 3.33 & 3.42 & 0.00 & 13.39 & 41.85 & 13.38 & 36.59 & 14.34 \\
    \midrule
    
    \bluegroupheader{9}{(III) Evaluation with Agent-Post-Trained Base Model}
    
    \bluerow
    Kimi-K2-Thinking
    & \metric{3.33\textsuperscript{*}}{\sameann}
    & \metric{3.42\textsuperscript{*}}{\sameann}
    & \metric{0.00\textsuperscript{*}}{\sameann}
    & \metric{13.39\textsuperscript{*}}{\sameann}
    & \metric{19.48}{\downann{22.37}}
    & \metric{13.38\textsuperscript{$\dagger$}}{\sameann}
    & \metric{36.59\textsuperscript{*}}{\sameann}
    & \metric{12.24}{\downann{2.10}} \\
    
    \bluerow
    Qwen3-Max
    & \metric{0.00}{\downann{3.33}}
    & \metric{2.41}{\downann{1.01}}
    & \metric{0.00}{\sameann}
    & \metric{8.04}{\downann{5.35}}
    & \metric{42.61}{\upann{0.76}}
    & \metric{13.38\textsuperscript{*}}{\sameann}
    & \metric{46.34}{\upann{9.75}}
    & \metric{13.39}{\downann{0.95}} \\
    
    \bluerow
    MiniMax-M2.5
    & \metric{0.00}{\downann{3.33}}
    & \metric{3.42\textsuperscript{*}}{\sameann}
    & \metric{0.00}{\sameann}
    & \metric{13.39}{\sameann}
    & \metric{41.85\textsuperscript{*}}{\sameann}
    & \metric{16.03}{\upann{2.65}}
    & \metric{33.54}{\downann{3.05}}
    & \metric{13.75}{\downann{0.59}} \\
    
    \bluerow
    MiniMax-M2.1
    & \metric{3.33}{\sameann}
    & \metric{3.42\textsuperscript{*}}{\sameann}
    & \metric{0.00}{\sameann}
    & \metric{10.49}{\downann{2.90}}
    & \metric{41.85\textsuperscript{$\dagger$}}{\sameann}
    & \metric{13.38\textsuperscript{*}}{\sameann}
    & \metric{38.41}{\upann{1.82}}
    & \metric{13.88}{\downann{0.46}} \\
    
    \bluerow
    Sonnet-4.5
    & \metric{3.33}{\sameann}
    & \metric{3.42}{\sameann}
    & \metric{2.00}{\upann{2.00}}
    & \metric{13.39}{\sameann}
    & \metric{41.85}{\sameann}
    & \metric{9.13}{\downann{4.25}}
    & \metric{44.51}{\upann{7.92}}
    & \metric{14.54}{\upann{0.20}} \\
    
    \bluerow
    GPT-5.1-Codex Max 
    & \metric{0.00}{\downann{3.33}}
    & \metric{3.77\textsuperscript{*}}{\upann{0.35}}
    & \metric{30.00\textsuperscript{*}}{\upann{30.00}}
    & \metric{23.29\textsuperscript{*}}{\upann{9.90}}
    & \metric{70.13\textsuperscript{*}}{\upann{28.28}}
    & \metric{12.21}{\downann{1.17}}
    & \metric{41.26}{\upann{4.67}}
    & \metric{21.02}{\upann{6.68}} \\
    
    \bluerow
    Gemini-3.1-Pro
    & \metric{2.22}{\downann{1.11}}
    & \metric{4.98}{\upann{1.56}}
    & \metric{57.33}{\upann{57.33}}
    & \metric{20.16}{\upann{6.77}}
    & \metric{47.54}{\upann{5.69}}
    & \metric{8.82}{\downann{4.56}}
    & \metric{52.03}{\upann{15.44}}
    & \metric{21.35}{\upann{7.01}} \\
    
    \bluerow
    GPT-5.2
    & \metric{1.11}{\downann{2.22}}
    & \metric{5.47}{\upann{2.05}}
    & \metric{58.33}{\upann{58.33}}
    & \metric{23.07}{\upann{9.68}}
    & \metric{64.59}{\upann{22.74}}
    & \metric{11.79}{\downann{1.59}}
    & \metric{37.40}{\upann{0.81}}
    & \metric{22.47}{\upann{8.13}} \\
    
    \bluerow
    Opus-4.6
    & \metric{5.56}{\upann{2.23}}
    & \metric{6.34}{\upann{2.92}}
    & \metric{96.67}{\upann{96.67}}
    & \metric{25.30}{\upann{11.91}}
    & \metric{52.19}{\upann{10.34}}
    & \metric{22.87}{\upann{9.49}}
    & \metric{36.59}{\sameann}
    & \metric{27.71}{\upann{13.37}} \\
    
    \bluerow
    Opus-4.7 (xHigh)
    & \metric{\second{7.78}}{\upann{4.45}}
    & \metric{21.90}{\upann{18.48}}
    & \metric{62.00}{\upann{62.00}}
    & \metric{24.03}{\upann{10.64}}
    & \metric{63.76}{\upann{21.91}}
    & \metric{16.53}{\upann{3.15}}
    & \metric{62.60}{\upann{26.01}}
    & \metric{30.38}{\upann{16.04}} \\
    
    \bluerow
    Opus-4.6 (1M)
    & \metric{6.67}{\upann{3.34}}
    & \metric{5.55}{\upann{2.13}}
    & \metric{\second{97.44}}{\upann{97.44}}
    & \metric{29.91}{\upann{16.52}}
    & \metric{78.75}{\upann{36.90}}
    & \metric{10.71}{\downann{2.67}}
    & \metric{57.93}{\upann{21.34}}
    & \metric{31.48}{\upann{17.14}} \\
    
    \bluerow
    GPT-5.4 (High)
    & \metric{6.67}{\upann{3.34}}
    & \metric{\second{49.53}}{\upann{46.11}}
    & \metric{\best{100.00}}{\upann{100.00}}
    & \metric{\best{34.15}}{\upann{20.76}}
    & \metric{\best{83.47}}{\upann{41.62}}
    & \metric{\second{29.41}}{\upann{16.03}}
    & \metric{\best{66.46}}{\upann{29.87}}
    & \metric{\best{41.40}}{\upann{27.06}} \\
    
    \midrule
    
    \yellowgroupheader{9}{(IV) Evaluation with Proposed Method (Compared with Baseline)}
    
    \lightyellowrowstart
    \textbf{GLM-4.7 (OpenCode)}
    & \metric{3.33\textsuperscript{*}}{\sameann}
    & \metric{3.42\textsuperscript{*}}{\sameann}
    & \metric{0.00\textsuperscript{*}}{\sameann}
    & \metric{13.39\textsuperscript{*}}{\sameann}
    & \metric{45.72}{\upann{3.87}}
    & \metric{13.38\textsuperscript{*}}{\sameann}
    & \metric{36.59\textsuperscript{*}}{\sameann}
    & \metric{14.70}{\upann{0.36}} \\

    \lightyellowrowstart
    \textbf{GLM-4.7 (Scaffold-only)}
    & \metric{0.00}{\downann{3.33}}
    & \metric{34.41}{\upann{30.99}}
    & \metric{63.00}{\upann{63.00}}
    & \metric{22.10}{\upann{8.71}}
    & \metric{67.55}{\upann{25.70}}
    & \metric{9.71}{\downann{3.67}}
    & \metric{56.70}{\upann{20.11}}
    & \metric{26.90}{\upann{12.56}} \\

    \rowcolor{oursblue}
    \textbf{GLM-4.7 \& ANDES (Ours)}
    & \metric{\best{9.00}}{\upann{5.67}}
    & \metric{\best{50.43}}{\upann{47.01}}
    & \metric{90.00}{\upann{90.00}}
    & \metric{\second{30.21}}{\upann{16.82}}
    & \metric{\second{82.17}}{\upann{40.32}}
    & \metric{\best{29.69}}{\upann{16.31}}
    & \metric{\second{65.35}}{\upann{28.76}}
    & \metric{\second{40.18}}{\upann{25.84}} \\
    
    \bottomrule
    \end{tabular}
    \end{adjustbox}
    
    \vspace{0.35em}
    {\footnotesize
    \textsuperscript{*} Model not submitted -- base model score shown
    \hspace{1.5em}
    \textsuperscript{$\dagger$} Evaluation error -- base model score shown
    }
    \end{table*}

\clearpage
\begingroup
\setcounter{topnumber}{5}
\setcounter{bottomnumber}{5}
\setcounter{totalnumber}{6}
\setcounter{dbltopnumber}{5}

\renewcommand{\topfraction}{0.98}
\renewcommand{\bottomfraction}{0.98}
\renewcommand{\textfraction}{0.01}
\renewcommand{\floatpagefraction}{0.01}
\renewcommand{\dbltopfraction}{0.98}
\renewcommand{\dblfloatpagefraction}{0.01}

\setlength{\floatsep}{3pt plus 1pt minus 1pt}
\setlength{\textfloatsep}{3pt plus 1pt minus 1pt}
\setlength{\intextsep}{3pt plus 1pt minus 1pt}
\setlength{\dblfloatsep}{3pt plus 1pt minus 1pt}
\setlength{\dbltextfloatsep}{3pt plus 1pt minus 1pt}

\begin{table*}[t]
    \centering
    \setlength{\tabcolsep}{3.9pt}
    \renewcommand{\arraystretch}{1.18}
    \caption{Experiments results on PostTrainBench (accuracy in \%) based on the SmolLM3-3B base model. The best and second-best results are highlighted in \textbf{bold} and \underline{underlined}. Numbers after arrows denote absolute changes over the corresponding zero-shot base model, with improvements in \textcolor{myred}{red} and declines in \textcolor{mygreen}{green}.}
    \vspace{4pt}
    \label{tab:smollm3-3b}
    
    \begin{adjustbox}{width=\textwidth,center}
    \fontsize{12.3}{14.2}\selectfont
    \begin{tabular}{l|cccccccc}
    \toprule
    \addlinespace[0.12em]
    \textbf{Method} & \textbf{AIME 2025} & \textbf{ArenaHard} & \textbf{BFCL} & \textbf{GPQA Main} & \textbf{GSM8K} & \textbf{HealthBench} & \textbf{HumanEval} & \textbf{AVG (\%)} \\[0.16em]
    \midrule
    
    \groupheader{9}{(I) Evaluation with Official Instruct Model}
    \grouprow
    \textbf{Official Instruct Model}
    & 26.67 & 49.20 & 84.00 & 33.26 & 82.18 & 29.58 & 70.12 & 44.81 \\
    \midrule
    
    \groupheader{9}{(II) Evaluation with Base Model (Zero-Shot)}
    \grouprow
    \textbf{Base Model (SmolLM3-3B)}
    & 3.33 & 0.42 & 0.00 & 4.91 & 21.08 & 0.00 & 6.10 & 4.52 \\
    \midrule
    
    \bluegroupheader{9}{(III) Evaluation with Agent-Post-Trained Base Model}
    
    \bluerow
    Sonnet-4.5
    & \metric{0.00}{\downann{3.33}}
    & \metric{0.24}{\downann{0.18}}
    & \metric{0.00}{\sameann}
    & \metric{3.21}{\downann{1.70}}
    & \metric{21.08}{\sameann}
    & \metric{0.00}{\sameann}
    & \metric{12.20}{\upann{6.10}}
    & \metric{3.99}{\downann{0.53}} \\
    
    \bluerow
    Qwen3-Max
    & \metric{3.33}{\sameann}
    & \metric{0.21}{\downann{0.21}}
    & \metric{0.00}{\sameann}
    & \metric{4.91}{\sameann}
    & \metric{21.08}{\sameann}
    & \metric{0.00\textsuperscript{*}}{\sameann}
    & \metric{6.10}{\sameann}
    & \metric{4.50}{\downann{0.02}} \\
    
    \bluerow
    Kimi-K2-Thinking
    & \metric{3.33\textsuperscript{*}}{\sameann}
    & \metric{0.42\textsuperscript{*}}{\sameann}
    & \metric{0.00\textsuperscript{*}}{\sameann}
    & \metric{4.91\textsuperscript{*}}{\sameann}
    & \metric{21.08}{\sameann}
    & \metric{0.00\textsuperscript{$\dagger$}}{\sameann}
    & \metric{6.10\textsuperscript{*}}{\sameann}
    & \metric{4.52}{\sameann} \\
    
    \bluerow
    MiniMax-M2.1
    & \metric{0.00}{\downann{3.33}}
    & \metric{0.42\textsuperscript{*}}{\sameann}
    & \metric{0.00}{\sameann}
    & \metric{12.05}{\upann{7.14}}
    & \metric{21.08\textsuperscript{$\dagger$}}{\sameann}
    & \metric{0.00\textsuperscript{*}}{\sameann}
    & \metric{5.49}{\downann{0.61}}
    & \metric{5.30}{\upann{0.78}} \\
    
    \bluerow
    GPT-5.1-Codex-Max 
    & \metric{3.33}{\sameann}
    & \metric{0.42\textsuperscript{*}}{\sameann}
    & \metric{0.00\textsuperscript{*}}{\sameann}
    & \metric{21.65\textsuperscript{*}}{\upann{16.74}}
    & \metric{21.08\textsuperscript{*}}{\sameann}
    & \metric{0.00}{\sameann}
    & \metric{5.49}{\downann{0.61}}
    & \metric{8.21}{\upann{3.69}} \\
    
    \bluerow
    MiniMax-M2.5
    & \metric{0.00}{\downann{3.33}}
    & \metric{2.18\textsuperscript{*}}{\upann{1.76}}
    & \metric{0.00}{\sameann}
    & \metric{17.19}{\upann{12.28}}
    & \metric{33.13\textsuperscript{*}}{\upann{12.05}}
    & \metric{0.00}{\sameann}
    & \metric{17.68}{\upann{11.58}}
    & \metric{9.04}{\upann{4.52}} \\
    
    \bluerow
    GPT-5.4 (High)
    & \metric{0.00}{\downann{3.33}}
    & \metric{\best{14.34}}{\upann{13.92}}
    & \metric{29.67}{\upann{29.67}}
    & \metric{\best{29.02}}{\upann{24.11}}
    & \metric{50.14}{\upann{29.06}}
    & \metric{18.64}{\upann{18.64}}
    & \metric{24.19}{\upann{18.09}}
    & \metric{20.72}{\upann{16.20}} \\
    
    \bluerow
    GPT-5.2
    & \metric{0.00}{\downann{3.33}}
    & \metric{6.49}{\upann{6.07}}
    & \metric{33.33}{\upann{33.33}}
    & \metric{25.82}{\upann{20.91}}
    & \metric{56.08}{\upann{35.00}}
    & \metric{\second{21.92}}{\upann{21.92}}
    & \metric{29.27}{\upann{23.17}}
    & \metric{21.26}{\upann{16.74}} \\
    
    \bluerow
    Gemini-3.1-Pro
    & \metric{\second{12.22}}{\upann{8.89}}
    & \metric{6.42}{\upann{6.00}}
    & \metric{27.67}{\upann{27.67}}
    & \metric{18.01}{\upann{13.10}}
    & \metric{55.42}{\upann{34.34}}
    & \metric{18.45}{\upann{18.45}}
    & \metric{\second{38.01}}{\upann{31.91}}
    & \metric{22.08}{\upann{17.56}} \\
    
    \bluerow
    Opus-4.6
    & \metric{5.00}{\upann{1.67}}
    & \metric{7.78}{\upann{7.36}}
    & \metric{75.92}{\upann{75.92}}
    & \metric{25.52}{\upann{20.61}}
    & \metric{41.04}{\upann{19.96}}
    & \metric{18.81}{\upann{18.81}}
    & \metric{24.75}{\upann{18.65}}
    & \metric{23.16}{\upann{18.64}} \\
    
    \bluerow
    Opus-4.7 (xHigh)
    & \metric{10.00}{\upann{6.67}}
    & \metric{10.21}{\upann{9.79}}
    & \metric{62.33}{\upann{62.33}}
    & \metric{\second{27.83}}{\upann{22.92}}
    & \metric{\second{66.11}}{\upann{45.03}}
    & \metric{16.53}{\upann{16.53}}
    & \metric{35.57}{\upann{29.47}}
    & \metric{25.28}{\upann{20.76}} \\
    
    \bluerow
    Opus-4.6 (1M)
    & \metric{\best{14.44}}{\upann{11.11}}
    & \metric{9.00}{\upann{8.58}}
    & \metric{\best{86.67}}{\upann{86.67}}
    & \metric{26.41}{\upann{21.50}}
    & \metric{58.05}{\upann{36.97}}
    & \metric{21.12}{\upann{21.12}}
    & \metric{25.61}{\upann{19.51}}
    & \metric{\second{28.52}}{\upann{24.00}} \\
    
    \midrule
    
    \yellowgroupheader{9}{(IV) Evaluation with Proposed Method (Compared with Baseline)}
    
    \lightyellowrowstart
    \textbf{GLM-4.7 (OpenCode)}
    & \metric{3.33\textsuperscript{*}}{\sameann}
    & \metric{0.42\textsuperscript{*}}{\sameann}
    & \metric{0.00}{\sameann}
    & \metric{4.91\textsuperscript{*}}{\sameann}
    & \metric{10.54}{\downann{10.54}}
    & \metric{0.00\textsuperscript{*}}{\sameann}
    & \metric{6.10\textsuperscript{*}}{\sameann}
    & \metric{3.53}{\downann{0.99}} \\

    \lightyellowrowstart
    \textbf{GLM-4.7 (Scaffold-only)}
    & \metric{0.00}{\downann{3.33}}
    & \metric{17.32}{\upann{16.90}}
    & \metric{70.00}{\upann{70.00}}
    & \metric{19.31}{\upann{14.40}}
    & \metric{48.36}{\upann{27.28}}
    & \metric{7.38}{\upann{7.38}}
    & \metric{29.27}{\upann{23.17}}
    & \metric{20.12}{\upann{15.60}} \\

    \rowcolor{oursblue}
    \textbf{GLM-4.7 \& ANDES (Ours)}
    & \metric{6.60}{\upann{3.27}}
    & \metric{\second{14.11}}{\upann{13.69}}
    & \metric{\second{78.67}}{\upann{78.67}}
    & \metric{25.80}{\upann{20.89}}
    & \metric{\best{67.52}}{\upann{46.44}}
    & \metric{\best{37.20}}{\upann{37.20}}
    & \metric{\best{41.37}}{\upann{35.27}}
    & \metric{\best{31.91}}{\upann{27.39}} \\
    
    \bottomrule
    \end{tabular}
    \end{adjustbox}
    
    \vspace{0.35em}
    {\footnotesize
    \textsuperscript{*} Model not submitted -- base model score shown
    \hspace{1.5em}
    \textsuperscript{$\dagger$} Evaluation error -- base model score shown
    }
    \end{table*}
\begin{table*}[t]
    \centering
    \setlength{\tabcolsep}{3.9pt}
    \renewcommand{\arraystretch}{1.18}
    \caption{Experiments results on PostTrainBench (accuracy in \%) based on the Gemma3-4B base model. The best and second-best results are highlighted in \textbf{bold} and \underline{underlined}. Numbers after arrows denote absolute changes over the corresponding zero-shot base model, with improvements in \textcolor{myred}{red} and declines in \textcolor{mygreen}{green}.}
    \vspace{4pt}
    \label{tab:gemma3-4b}
    
    \begin{adjustbox}{width=\textwidth,center}
    \fontsize{12.3}{14.2}\selectfont
    \begin{tabular}{l|cccccccc}
    \toprule
    \addlinespace[0.12em]
    \textbf{Method} & \textbf{AIME 2025} & \textbf{ArenaHard} & \textbf{BFCL} & \textbf{GPQA Main} & \textbf{GSM8K} & \textbf{HealthBench} & \textbf{HumanEval} & \textbf{AVG (\%)} \\[0.16em]
    \midrule
    
    \groupheader{9}{(I) Evaluation with Official Instruct Model}
    \grouprow
    \textbf{Official Instruct Model}
    & 10.00 & 94.80 & 67.00 & 31.47 & 83.55 & 46.06 & 69.51 & 46.58 \\
    \midrule
    
    \groupheader{9}{(II) Evaluation with Base Model (Zero-Shot)}
    \grouprow
    \textbf{Base Model (Gemma3-4B)}
    & 0.00 & 0.29 & 6.00 & 1.56 & 6.14 & 17.04 & 0.61 & 4.60 \\
    \midrule
    
    \bluegroupheader{9}{(III) Evaluation with Agent-Post-Trained Base Model}
    
    \bluerow
    Kimi-K2-Thinking
    & \metric{0.00\textsuperscript{*}}{\sameann}
    & \metric{0.29\textsuperscript{*}}{\sameann}
    & \metric{6.00\textsuperscript{*}}{\sameann}
    & \metric{1.56\textsuperscript{*}}{\sameann}
    & \metric{6.14}{\sameann}
    & \metric{17.04\textsuperscript{$\dagger$}}{\sameann}
    & \metric{0.61\textsuperscript{*}}{\sameann}
    & \metric{4.60}{\sameann} \\
    
    \bluerow
    Qwen3-Max
    & \metric{0.00}{\sameann}
    & \metric{0.29}{\sameann}
    & \metric{6.00}{\sameann}
    & \metric{1.56}{\sameann}
    & \metric{6.14}{\sameann}
    & \metric{17.04\textsuperscript{*}}{\sameann}
    & \metric{0.61}{\sameann}
    & \metric{4.60}{\sameann} \\
    
    \bluerow
    MiniMax-M2.5
    & \metric{0.00}{\sameann}
    & \metric{3.60\textsuperscript{*}}{\upann{3.31}}
    & \metric{9.00}{\upann{3.00}}
    & \metric{1.56}{\sameann}
    & \metric{36.39\textsuperscript{*}}{\upann{30.25}}
    & \metric{18.34}{\upann{1.30}}
    & \metric{3.05}{\upann{2.44}}
    & \metric{8.45}{\upann{3.85}} \\
    
    \bluerow
    MiniMax-M2.1
    & \metric{0.00}{\sameann}
    & \metric{0.29\textsuperscript{*}}{\sameann}
    & \metric{54.00}{\upann{48.00}}
    & \metric{1.56}{\sameann}
    & \metric{2.96\textsuperscript{$\dagger$}}{\downann{3.18}}
    & \metric{17.04\textsuperscript{*}}{\sameann}
    & \metric{30.49}{\upann{29.88}}
    & \metric{11.06}{\upann{6.46}} \\
    
    \bluerow
    Sonnet-4.5
    & \metric{0.00}{\sameann}
    & \metric{0.29}{\sameann}
    & \metric{5.00}{\downann{1.00}}
    & \metric{25.22}{\upann{23.66}}
    & \metric{\second{57.92}}{\upann{51.78}}
    & \metric{10.69}{\downann{6.35}}
    & \metric{34.76}{\upann{34.15}}
    & \metric{17.14}{\upann{12.54}} \\
    
    \bluerow
    GPT-5.1-Codex Max 
    & \metric{0.00}{\sameann}
    & \metric{8.45\textsuperscript{*}}{\upann{8.16}}
    & \metric{33.67\textsuperscript{*}}{\upann{27.67}}
    & \metric{21.21\textsuperscript{*}}{\upann{19.65}}
    & \metric{44.00\textsuperscript{*}}{\upann{37.86}}
    & \metric{20.13}{\upann{3.09}}
    & \metric{33.54}{\upann{32.93}}
    & \metric{19.42}{\upann{14.82}} \\
    
    \bluerow
    Gemini-3.1-Pro
    & \metric{1.11}{\upann{1.11}}
    & \metric{16.01}{\upann{15.72}}
    & \metric{80.67}{\upann{74.67}}
    & \metric{19.79}{\upann{18.23}}
    & \metric{39.50}{\upann{33.36}}
    & \metric{19.70}{\upann{2.66}}
    & \metric{34.55}{\upann{33.94}}
    & \metric{23.15}{\upann{18.55}} \\
    
    \bluerow
    Opus-4.6 (1M)
    & \metric{0.00}{\sameann}
    & \metric{13.95}{\upann{13.66}}
    & \metric{64.33}{\upann{58.33}}
    & \metric{28.12}{\upann{26.56}}
    & \metric{36.42}{\upann{30.28}}
    & \metric{21.49}{\upann{4.45}}
    & \metric{\second{36.38}}{\upann{35.77}}
    & \metric{23.60}{\upann{19.00}} \\
    
    \bluerow
    GPT 5.4 (High)
    & \metric{3.33}{\upann{3.33}}
    & \metric{0.29}{\sameann}
    & \metric{\best{100.00}}{\upann{94.00}}
    & \metric{26.34}{\upann{24.78}}
    & \metric{44.20}{\upann{38.06}}
    & \metric{21.90}{\upann{4.86}}
    & \metric{23.78}{\upann{23.17}}
    & \metric{24.85}{\upann{20.25}} \\
    
    \bluerow
    GPT-5.2
    & \metric{0.00}{\sameann}
    & \metric{13.20}{\upann{12.91}}
    & \metric{89.00}{\upann{83.00}}
    & \metric{\second{28.57}}{\upann{27.01}}
    & \metric{51.91}{\upann{45.77}}
    & \metric{20.19}{\upann{3.15}}
    & \metric{21.54}{\upann{20.93}}
    & \metric{25.11}{\upann{20.51}} \\
    
    \bluerow
    Opus-4.6
    & \metric{\best{5.56}}{\upann{5.56}}
    & \metric{6.34}{\upann{6.05}}
    & \metric{\second{96.67}}{\upann{90.67}}
    & \metric{25.30}{\upann{23.74}}
    & \metric{52.19}{\upann{46.05}}
    & \metric{\second{22.87}}{\upann{5.83}}
    & \metric{\best{36.59}}{\upann{35.98}}
    & \metric{27.71}{\upann{23.11}} \\
    
    \bluerow
    Opus-4.7 (xHigh)
    & \metric{1.11}{\upann{1.11}}
    & \metric{\best{30.86}}{\upann{30.57}}
    & \metric{91.33}{\upann{85.33}}
    & \metric{\best{28.72}}{\upann{27.16}}
    & \metric{55.70}{\upann{49.56}}
    & \metric{21.49}{\upann{4.45}}
    & \metric{27.85}{\upann{27.24}}
    & \metric{\second{28.43}}{\upann{23.83}} \\
    
    \midrule
    
    \yellowgroupheader{9}{(IV) Evaluation with Proposed Method (Compared with Baseline)}
    
    \lightyellowrowstart
    \textbf{GLM-4.7 (OpenCode)}
    & \metric{\second{4.60}}{\upann{4.60}}
    & \metric{0.00\textsuperscript{$\dagger$}}{\downann{0.29}}
    & \metric{0.29\textsuperscript{*}}{\downann{5.71}}
    & \metric{6.00\textsuperscript{$\dagger$}}{\upann{4.44}}
    & \metric{1.56\textsuperscript{$\dagger$}}{\downann{4.58}}
    & \metric{6.14\textsuperscript{$\dagger$}}{\downann{10.90}}
    & \metric{17.04\textsuperscript{*}}{\upann{16.43}}
    & \metric{0.61\textsuperscript{$\dagger$}}{\downann{3.99}} \\

    \lightyellowrowstart
    \textbf{GLM-4.7 (Scaffold-only)}
    & \metric{0.00}{\sameann}
    & \metric{31.12}{\upann{30.83}}
    & \metric{61.00}{\upann{55.00}}
    & \metric{22.76}{\upann{21.20}}
    & \metric{36.16}{\upann{30.02}}
    & \metric{4.89}{\downann{12.15}}
    & \metric{28.29}{\upann{27.68}}
    & \metric{19.76}{\upann{15.16}} \\

    \rowcolor{oursblue}
    \textbf{GLM-4.7 \& ANDES (Ours)}
    & \metric{0.00}{\sameann}
    & \metric{\second{25.16}}{\upann{24.87}}
    & \metric{85.00}{\upann{79.00}}
    & \metric{25.42}{\upann{23.86}}
    & \metric{\best{76.57}}{\upann{70.43}}
    & \metric{\best{33.28}}{\upann{16.24}}
    & \metric{32.11}{\upann{31.50}}
    & \metric{\best{31.02}}{\upann{26.42}} \\
    
    \bottomrule
    \end{tabular}
    \end{adjustbox}
    
    \vspace{0.35em}
    {\footnotesize
    \textsuperscript{*} Model not submitted -- base model score shown
    \hspace{1.5em}
    \textsuperscript{$\dagger$} Evaluation error -- base model score shown
    }
    \end{table*}

\clearpage
\endgroup

\section{Specific Composition of World Tree}
\label{world_tree}

\begin{figure}[t]
    \centering
    \includegraphics[width=\linewidth]{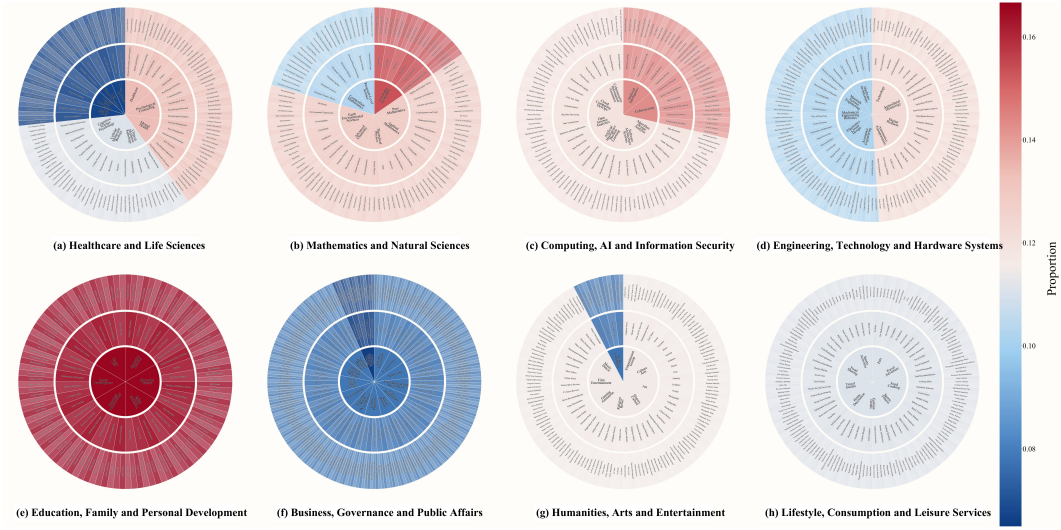}
    \caption{\textbf{Visualization of the knowledge coverage of the \textsc{Andes} world tree. } It shows the hierarchical distribution of all world-tree nodes across broad macro domains, covering diverse topics, themes, and scenarios. }
    \label{fig:world_tree_8category}
\end{figure}

Fig.~\ref{fig:world_tree_8category} first shows the knowledge distribution of all nodes in the world tree. 
The initialized world tree spans broad macro categories, including healthcare and life sciences, mathematics and natural sciences, computing and information security, engineering and technology, education and personal development, business and governance, humanities and arts, and lifestyle services. 
Each category contains rich topic, theme, and scenario nodes, forming a broad and hierarchical synthesis space. 
Therefore, \textsc{Andes} does not rely on a single domain or fixed templates for data generation. 
Instead, it can perform target-aware routing and dynamic expansion over a wide knowledge space, which supports the diversity, generality, and cross-benchmark adaptability of the synthesized data.

\section{Distribution of Synthesized Questions in the Embedding Space}
\label{world_tree_tsne}

\begin{figure}[ht]
    \centering
    \includegraphics[width=0.45\linewidth]{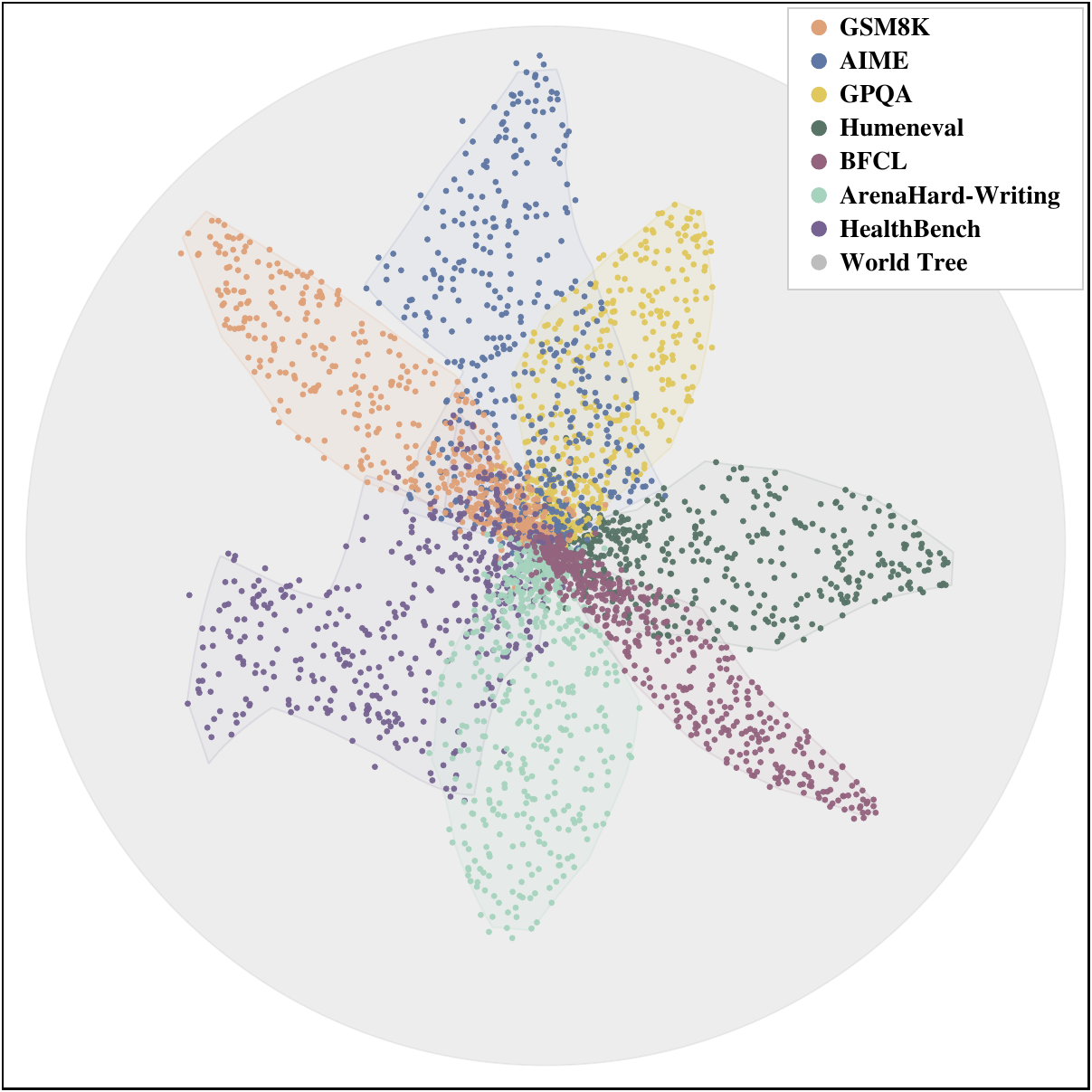}
    \caption{\textbf{t-sne visualization of the question of the \textsc{Andes} world tree.}
    It shows that the world tree provides a broad and general knowledge space that supports diverse target tasks.}
    \label{fig:world_tree_tsne}
\end{figure}

Fig.~\ref{fig:world_tree_tsne} visualizes the question embeddings of synthetic data generated for different PostTrainBench benchmarks. 
The samples from different benchmarks form distinguishable regions in the embedding space while still sharing overlapping areas near the center. 
This indicates that the world tree of \textsc{Andes} covers diverse capability directions, including mathematical reasoning, code generation, tool use, health dialogue, and open-ended writing. 
Meanwhile, the benchmark-specific distributions show that the directed routing mechanism can select suitable topics, themes, and scenarios according to the target task, thereby generating data aligned with the required capabilities.

\section{Case Study: Agent Trajectory}
\label{sec:case_study}

In this section, we present a representative agent trajectory of the trainer agent utilizing the \textsc{Andes} skill across multiple benchmarks. As illustrated below, the agent does not merely overfit to the surface-level patterns or specific formatting of individual benchmarks. Instead, guided by the \textsc{Andes} skill, the agent autonomously reconstructs the target requirements into a set of refined macro-cognitive domains.

Specifically, the agent decomposes complex benchmark prompts into fundamental operational capabilities, such as ``Quantitative Modeling and Relation Extraction'' for mathematics or ``Edge Cases and Robust Programming'' for code synthesis. By reorganizing task-local prompts into these macro domains, \textsc{Andes} ensures that the synthesized data captures the underlying reasoning logic rather than memorizing task-specific shortcuts. This hierarchical abstraction allows the agent to maintain high fidelity to the target benchmarks while simultaneously fostering robust cross-task generalization, effectively preventing diversity collapse during long-horizon autonomous alignment.

\definecolor{andesGray}{HTML}{77736C}
\definecolor{andesBorder}{HTML}{8F918F}
\definecolor{setupBack}{HTML}{FFF4DF}
\definecolor{domainBack}{HTML}{EAF4FF}
\definecolor{algoBack}{HTML}{F2EDFF}
\definecolor{trajBack}{HTML}{EDF8F1}
\definecolor{caseBack}{HTML}{F7F8FB}
\definecolor{roundBackOne}{HTML}{F8F8F8}
\definecolor{roundBackTwo}{HTML}{F2F2F2}
\definecolor{roundBackThree}{HTML}{EDEDED}
\definecolor{andesBlue}{HTML}{2468AD}
\definecolor{andesTeal}{HTML}{148F8B}
\definecolor{andesRed}{HTML}{BD2934}
\definecolor{andesGreen}{HTML}{2F8F5B}
\definecolor{andesOrange}{HTML}{CC7B12}
\definecolor{andesPurple}{HTML}{7259BF}

\tcbset{
  andesbox/.style={
    enhanced,
    breakable,
    arc=2mm,
    boxrule=0.7pt,
    colframe=andesBorder,
    coltitle=white,
    colbacktitle=andesGray,
    fonttitle=\bfseries\large,
    lefttitle=5mm,
    righttitle=5mm,
    toptitle=1.6mm,
    bottomtitle=1.6mm,
    titlerule=0pt,
    left=7mm,
    right=7mm,
    top=6mm,
    bottom=6mm,
    before skip=10pt,
    after skip=12pt
  },
  interactionround/.style={
    enhanced,
    arc=1.5mm,
    boxrule=0.45pt,
    colframe=andesBorder,
    colback=roundBackOne,
    coltitle=white,
    colbacktitle=andesGray,
    fonttitle=\bfseries\scriptsize,
    attach boxed title to top left={xshift=3mm,yshift=-2mm},
    boxed title style={
      enhanced,
      arc=1mm,
      boxrule=0pt,
      colframe=andesGray,
      colback=andesGray,
      left=2.5mm,
      right=2.5mm,
      top=0.6mm,
      bottom=0.6mm
    },
    borderline west={1.6pt}{0pt}{andesGray},
    left=4mm,
    right=4mm,
    top=5mm,
    bottom=3mm,
    before skip=9pt,
    after skip=9pt,
    fontupper=\scriptsize
  }
}

\setlength{\parindent}{0pt}
\setlength{\parskip}{4pt}
\setlist[itemize]{leftmargin=1.25em,itemsep=2pt,topsep=2pt}
\setlist[enumerate]{leftmargin=1.35em,itemsep=3pt,topsep=2pt}
\sloppy
\emergencystretch=2em

\subsection*{F1. Initial Input and To-Do List}
This module defines the run contract before any synthesis begins: the target benchmark, the base model, the hard constraints, and the execution checklist. It anchors the rest of the trajectory so the reader can interpret later decisions as part of one controlled GSM8K post-training run.

\begin{tcolorbox}[andesbox,colback=setupBack,title={F1. Initial Input and To-Do List}]
\textbf{Initial input.}

\textbf{(a)} Target benchmark: GSM8K;\par\smallskip
\textbf{(b)} Base model: Qwen3-4B-Base;\par\smallskip
\textbf{(c)} Objective: Fine-tune base model for grade-school arithmetic word-problem reasoning;\par\medskip

\textbf{Hard constraints.}

\textbf{(a)} Finish within 10 hours;\par\smallskip
\textbf{(b)} Fine-tune the base model only;\par\smallskip
\textbf{(c)} Avoid GSM8K test leakage;\par\smallskip
\textbf{(d)} Synthesize data through \textsc{Andes} only;\par\smallskip
\textbf{(e)} Retain  $\sim$8,000 Alpaca-format samples;\par\medskip

\textbf{Execution to-do list.}

\textbf{(a)} Analyze GSM8K requirements and divide them into cognitive-operation domains;\par\smallskip
\textbf{(b)} Use \textsc{Andes} to synthesize domain-targeted training data;\par\smallskip
\textbf{(c)} Merge retained samples into Alpaca instruction/input/output format;\par\smallskip
\textbf{(d)} Run full SFT with LlamaFactory on Qwen3-4B-Base;\par\smallskip
\textbf{(e)} Evaluate the best checkpoint and save it as final model;
\end{tcolorbox}

\subsection*{F2. Capability-Domain Agent Trajectory}
\vspace{-6pt}
This module merges the macro capability-domain split with the full interaction trajectory. The domain split jointly considers the model abilities that need strengthening, the reasoning demands expressed by the benchmark content, and the post-training coverage needed for robust generalization. The agent then turns these domains into synthesis specifications, delegates generation to \textsc{Andes}, reads reports as feedback signals, and uses each report to control retention and sharpen the next task description. This report-driven loop connects data synthesis, quality control, aggregation, training, and evaluation into a complete automated pipeline for high-quality post-training.
\vspace{-4pt}

\begin{tcolorbox}[andesbox,colback=trajBack,title={F2. Capability-Domain Agent Trajectory}]
\small

\textbf{Boot the agent shell.}
(a) The trainer agent loads the post-training, data-synthesis, and training-configuration skills;
(b) Parses the benchmark instruction and resource constraints from the task prompt;
(c) Establishes a sandboxed execution plan with synthesis, training, and evaluation phases.

\par\smallskip
\textbf{Research the benchmark.}
(a) The agent inspects GSM8K-style reasoning requirements rather than copying benchmark examples;
(b) Identifies the core capability demand: arithmetic word-problem solving with chained intermediate states;
(c) Converts surface problem formats into abstract cognitive operations such as extraction, modeling, constraint checking, and answer verification.

\par\smallskip
\textbf{Construct the domain plan.}
(a) The agent partitions the benchmark into non-overlapping macro capability domains;
(b) Each domain is defined by reasoning structure rather than topic keywords;
(c) The plan is checked for coverage balance so that no single arithmetic pattern dominates later synthesis.

\par\smallskip
\textbf{Author the synthesis specification.}
(a) For each domain, the agent writes a task description \(\tau_j\) that states the target capability, difficulty style, and constraint structure;
(b) It assigns a synthesis budget \(N_j\) and a fixed format protocol \(\phi_j\);
(c) It records expected failure modes, including overly narrow descriptions, template lock-in, constraint drift, and shallow form-only reasoning.

\par\smallskip
\textbf{Delegate synthesis to \textsc{Andes}.}
(a) The agent invokes \textsc{Andes} as an external synthesis tool rather than as a planner;
(b) \textsc{Andes} routes each domain through the world-tree hierarchy from topic to theme to scenario;
(c) Fusion-track samples inject the domain requirement into diverse scenarios, while generic-track samples preserve broad contextual coverage.

\par\smallskip
\textbf{Read reports and decide retention.}
(a) After each domain run, the agent reads the synthesis report together with the original task description;
(b) It compares topic coverage, effective quantity, fusion/generic balance, logical diversity, and reported signals;
(c) It chooses a predefined random-discard tier only after this report--description comparison, avoiding any manual semantic filtering.

\textit{The following Round1--Round3 blocks are fully expanded examples of this interaction paradigm. Each round shows how one task description is instantiated into a synthesized case, how the \textsc{Andes} report exposes quality signals, and how the agent converts that report into a random-discard decision. The rounds are interactive rather than independent: the report from one round informs how the agent sharpens the next round's task description, coverage emphasis, and expected failure modes. The final ``Later Rounds'' block summarizes the remaining task-description/domain pairs, with ellipses marking omitted case-level examples.}

\par\smallskip
\begin{tcolorbox}[interactionround,title={Round1}]
\textbf{Task description.} Generate GSM8K-style arithmetic tasks that require a solver to maintain a running state across several dependent commercial operations, including discounts, giveaway rules, production costs, and final profit computation. The task should reward ordered reasoning rather than isolated arithmetic.\\
\textbf{Domain1:} \textbf{Multi-step Sequential Reasoning:}\ Generate arithmetic word problems that require 3 to 7 ordered operations. Each later operation depends on an earlier intermediate result, so the solver must maintain a running state rather than execute isolated arithmetic.\\
\textbf{Full task.} A startup sells eco-friendly bottles. Each bottle costs \$8 to produce and normally sells for \$20. During a promotion, the first 100 sold bottles receive a 25\% discount. In addition, for every 5 bottles sold during the promotion, 1 extra bottle is given away for free. After the promotion, another 150 bottles are sold at the regular price. Compute total revenue and total profit.\\
\textbf{Capability focus.} The case requires a chained calculation: discounted price, promotional revenue, free-bottle count, production cost, regular-sale revenue, and final profit.\\
\textbf{Reasoning.} The discounted price is \$20 $\times$ (1 - 0.25) = \$15. Promotion revenue is 100 $\times$ \$15 = \$1,500. The free-bottle rule gives 100 / 5 = 20 free bottles, so the promotion requires producing 100 + 20 = 120 bottles, costing 120 $\times$ \$8 = \$960. After the promotion, regular revenue is 150 $\times$ \$20 = \$3,000, and regular production cost is 150 $\times$ \$8 = \$1,200. Total revenue is \$1,500 + \$3,000 = \$4,500, total cost is \$960 + \$1,200 = \$2,160, and profit is \$4,500 - \$2,160 = \$2,340.\\
\textbf{Answer.} Total revenue is \$4,500, and total profit is \$2,340.\\
\textbf{Report.} The synthesis report shows strong coverage of sequential arithmetic in commerce, promotion, and inventory scenarios. Most samples preserve a clear dependency chain from intermediate quantities to final profit, and the fusion track adds proportional details such as discounts and bonus units without overwhelming the D1 target. No severe coverage collapse is detected, but a small cluster of examples repeats the revenue--cost--profit template with similar ordering.\\
\textbf{Agent report analysis and retention decision.} The agent compares the report with the original task description and finds the intended multi-step dependency structure intact. Because the only material risk is mild template repetition rather than semantic drift, the agent chooses a 10\% random discard ratio and carries the anti-template warning into the next round's task description without performing content-based filtering.
\end{tcolorbox}

\begin{center}
{\large\textcolor{andesGray}{\(\Downarrow\)}}
\end{center}

\begin{tcolorbox}[interactionround,title={Round2}]
\textbf{Task description.} Generate GSM8K-style tasks where the solver must extract the relevant quantities from natural language, ignore distractor objects, align comparable categories, and aggregate matched, overpacked, or underpacked counts. The task should test modeling from text before calculation.\\
\textbf{Domain2:} \textbf{Information Extraction and Mathematical Modeling:}\ Extract multiple quantities and relationships from natural language, ignore distractors, and translate the remaining evidence into equations. The same surface pattern may map to different mathematical structures depending on context.\\
\textbf{Full task.} A travel guide recommends packing 10 shirts, 5 pairs of pants, 2 jackets, 8 pairs of socks, and 3 pairs of shoes. The traveler also packs 2 hats, 1 scarf, 4 books, and 1 camera. The actual relevant items are 12 shirts, 6 pairs of pants, 1 jacket, 10 pairs of socks, and 2 pairs of shoes. Compute correctly packed, overpacked, and underpacked items.\\
\textbf{Capability focus.} The case tests whether the model can ignore distractors and compare only the categories specified by the guide.\\
\textbf{Reasoning.} Shirts: 10 recommended and 12 packed, so 10 correct and 2 overpacked. Pants: 5 recommended and 6 packed, so 5 correct and 1 overpacked. Jackets: 2 recommended and 1 packed, so 1 correct and 1 underpacked. Socks: 8 recommended and 10 packed, so 8 correct and 2 overpacked. Shoes: 3 recommended and 2 packed, so 2 correct and 1 underpacked. Thus, correctly packed items are 10 + 5 + 1 + 8 + 2 = 26, overpacked items are 2 + 1 + 0 + 2 + 0 = 5, and underpacked items are 0 + 0 + 1 + 0 + 1 = 2. The extra items are 2 + 1 + 4 + 1 = 8 distractors and do not affect the requested totals.\\
\textbf{Answer.} Correctly packed: 26; overpacked: 5; underpacked: 2.\\
\textbf{Report.} The synthesis report indicates that D2 samples consistently require extraction before computation. Generated cases vary across travel, classroom, shopping, and household settings, while distractor quantities are usually separable from the categories requested by the task. The report finds no major template lock-in and no constraint drift; most samples keep the answer grounded in an explicit category comparison rather than an unstructured count of all mentioned objects.\\
\textbf{Agent report analysis and retention decision.} The agent reads the report against the task description and confirms that distractor suppression and mathematical modeling are the dominant operations. Since failure signals are low and the effective quantity is near the planned budget, the agent chooses a 0\% random discard ratio and uses the clean extraction signal to make the next task description emphasize explicit constraint verification.
\end{tcolorbox}

\begin{center}
{\large\textcolor{andesGray}{\(\Downarrow\)}}
\end{center}

\begin{tcolorbox}[interactionround,title={Round3}]
\textbf{Task description.} Generate GSM8K-style tasks requiring a feasibility or threshold judgment after arithmetic computation. The solver must compute intermediate quantities, compare them against an explicit constraint, and revise or verify the result when the original condition fails.\\
\textbf{Domain3:} \textbf{Constraint Satisfaction and Logical Reasoning:}\ Reason under explicit or implicit constraints: maximum or minimum values, budget limits, time limits, capacities, integer requirements, and interacting conditions. The answer must be feasible, not merely numerically computed.\\
\textbf{Full task.} A bookstore membership program gives members a 15\% discount. Members spend \$50 per visit on average. The non-discounted profit margin is 30\%. Compute the member profit per visit and verify whether the discounted margin remains at least 20\%. If it fails, suggest a revised discount.\\
\textbf{Capability focus.} The case requires proportional reasoning over revenue and cost, then a constraint check against a margin threshold and a verification of the revised discount.\\
\textbf{Reasoning.} The non-discounted cost is \$50 $\times$ (1 - 0.30) = \$35. With a 15\% discount, member revenue is \$50 $\times$ (1 - 0.15) = \$42.50, so profit is \$42.50 - \$35 = \$7.50. The resulting margin is \$7.50 / \$42.50 $\approx$ 17.65\%, which is below 20\%. Let the revised discount be \(x\). To meet the 20\% margin threshold, \((50(1-x)-35)/(50(1-x)) = 0.20\), which gives \(50 - 50x - 35 = 10 - 10x\), then \(5 = 40x\), so \(x = 0.125\). A 12.5\% discount gives revenue \$50 $\times$ 0.875 = \$43.75, profit \$43.75 - \$35 = \$8.75, and margin \$8.75 / \$43.75 = 20\%.\\
\textbf{Answer.} Under the 15\% discount, profit is \$7.50 and margin is about 17.65\%, so the constraint fails. A 12.5\% discount reaches the 20\% threshold; for strictly above 20\%, the discount should be slightly below 12.5\%.\\
\textbf{Report.} The synthesis report shows good coverage of budget, margin, capacity, and threshold checks, with many samples requiring the model to reject a numerically computed value when it violates a stated condition. However, the report also flags a moderate F5-style constraint-drift risk: some generated cases compute a revised value but do not explicitly verify that the revised value satisfies the original threshold. Logical diversity is acceptable, but the verification step is less consistently represented than the arithmetic step.\\
\textbf{Agent report analysis and retention decision.} The agent compares these findings with the D3 task description and treats verification omission as directly relevant to the target capability. Because the batch still contains useful constraint-reasoning data but has a moderate alignment risk, the agent chooses a 20\% random discard ratio while preserving the remaining records verbatim.
\end{tcolorbox}

\begin{center}
{\large\textcolor{andesGray}{\(\Downarrow\)}}
\end{center}

\begin{tcolorbox}[interactionround,title={Later Rounds}]
\textbf{Task description.} Generate GSM8K-style tasks that require ratios, rates, scaling, unit conversion, percentages, and quantity comparisons. The task should force the solver to preserve proportional relationships while moving across units or scales.\\
\textbf{Domain4:} \textbf{Proportional and Comparative Reasoning:}\ Handle ratios, rates, scaling, unit conversion, percentages, and comparisons between quantities. The key skill is preserving the proportional relationship while quantities move across units or scales.

\begin{center}
\textbf{......}
\end{center}

\textbf{Task description.} Generate GSM8K-style tasks that require decomposing a multi-component situation into ordered sub-problems, solving each piece, and recombining the partial results into one final answer.\\
\textbf{Domain5:} \textbf{Complex Problem Decomposition:}\ Break a multi-component task into 2 to 4 sub-problems, order the dependent pieces, solve them separately, and recombine partial answers. The domain targets planning structure as much as computation.

\begin{center}
\textbf{......}
\end{center}

\textbf{Task description.} Generate GSM8K-style tasks that require checking intermediate and final answers against the original constraints, detecting invalid outputs, and correcting the result when practical sanity conditions fail.\\
\textbf{Domain6:} \textbf{Verification and Self-Correction:}\ Check intermediate and final answers against the original constraints and practical sanity conditions. Examples include whole-number checks, capacity checks, budget checks, and detecting impossible configurations.

\begin{center}
\textbf{......}
\end{center}
\end{tcolorbox}

\par\smallskip
\textbf{Complete the synthesis loop.}
(a) The agent repeats the domain-level call, report reading, and discard-decision loop until all initial domains are completed;
(b) If the retained pool is still insufficient, it adds supplementary capability domains instead of resampling the same domain;
(c) Throughout the loop, the format protocol remains fixed and retained samples are not rewritten, deduplicated, or filtered by content.

\par\smallskip
\textbf{Aggregate training data.}
(a) The agent merges all retained domain outputs into one Alpaca-style dataset;
(b) Both fusion and generic records are preserved as valid training signals;
(c) The final dataset reflects the original capability split while maintaining scenario diversity from the world tree.

\par\smallskip
\textbf{Train and evaluate.}
(a) The agent launches full supervised fine-tuning on the selected base model;
(b) Training progress is monitored through logs and checkpoint artifacts;
(c) The final checkpoint is evaluated on the target benchmark to measure whether the synthesized curriculum improves reasoning performance.
\end{tcolorbox}




\end{document}